\documentclass{article}

\usepackage{arxiv}

\usepackage[authoryear,longnamesfirst]{natbib}

\usepackage{algorithm}
\usepackage[noend]{algpseudocode}
\usepackage{caption}
\usepackage{subcaption}
\usepackage{float}
\usepackage[normalem]{ulem}
\usepackage[most]{tcolorbox}
\usepackage{xcolor}
\usepackage{listings}
\usepackage{comment}
\usepackage{hyperref}
\usepackage{amssymb}
\usepackage{multirow}

\def\tsc#1{\csdef{#1}{\textsc{\lowercase{#1}}\xspace}}
\tsc{WGM}
\tsc{QE}
\tsc{EP}
\tsc{PMS}
\tsc{BEC}
\tsc{DE}

\newcommand{\Why}{why}
\newcommand{\Success}{\textit{Success}}
\newcommand{\Failure}{\textit{Failure}}
\newcommand{\Running}{\textit{Running}}
\newcommand{\Invalid}{\textit{Invalid}}

\newcommand{\twolinetablecell}[2][l]{%
  \begin{tabular}[#1]{@{}l@{}}#2\end{tabular}}

\makeatletter
\renewcommand{\Function}[2]{%
  \csname ALG@cmd@\ALG@L @Function\endcsname{#1}{#2}%
  \def\jayden@currentfunction{#1}%
}
\newcommand{\funclabel}[1]{%
  \@bsphack
  \protected@write\@auxout{}{%
    \string\newlabel{#1}{{\jayden@currentfunction}{\thepage}}%
  }%
  \@esphack
}
\makeatother

\newcommand*{\corresponds}{\ensuremath{~\widehat{=}~}}

\tcbset{
  reviewerdiagonal/.style={
    enhanced,
    frame hidden,
    colback=white,
    coltext=red,        % <-- this colors all text inside red
    left=0pt, right=0pt, top=4pt, bottom=4pt,
    overlay={%
      \begin{scope}
        \draw[red,line width=1pt]
          (frame.south west) -- (frame.north east);
        \draw[red,line width=1pt]
          (frame.north west) -- (frame.south east);
      \end{scope}
    }
  }
}

\newcommand{\ReviewerD}[1]{}
\excludecomment{ReviewerDSec}

\newcommand{\repourl}{\url{https://github.com/tamlinlove/btcm}}
\newcommand{\prompturl}{\url{https://github.com/tamlinlove/btcm/tree/main/btcm/experiment/example_prompts}}

\title{Temporal Counterfactual Explanations \\of Behaviour Tree Decisions}

\author{
 Tamlin Love, Antonio Andriella, Guillem Alenyà \\
  Institut de Robòtica i Informàtica Industrial, CSIC-UPC\\
  Llorens i Artigas 4-6\\
  Barcelona, Spain, 08028 \\
  \texttt{\{tlove,aandriella,galenya\}@iri.upc.edu} \\
}

\begin{document}
\maketitle
\begin{abstract}
Explainability, in particular, the ability for robots to explain why they have made a decision or behaved in a certain way, is a critical tool in helping users understand the robots they interact and coexist with. Behaviour trees are a popular framework for controlling the decision-making of robots, and thus a natural question to ask is whether or not a system driven by a behaviour tree is capable of answering ``why'' questions. While explainability for behaviour tree-driven robots has seen some prior attention, no existing methods are capable of generating causal, counterfactual explanations which detail the reasons for robot decisions and behaviour. Therefore, in this work, we introduce a novel approach which automatically generates counterfactual explanations in response to contrastive ``why'' questions. Our method achieves this by first automatically building a causal model from the structure of the behaviour tree as well as domain knowledge about the state and individual behaviour tree nodes. The resultant causal model is then queried and searched to find a set of diverse counterfactual explanations. We demonstrate that our approach is able to correctly explain the behaviour of a wide range of behaviour tree structures and states in real time, unlike previous methods which are either unable to answer contrastive questions with causal explanations, or are not guaranteed to provide consistent and accurate explanations. By being able to answer a wide range of causal queries, our approach represents a step towards more transparent, understandable, and ultimately safe and trustworthy robotic systems.
\end{abstract}

% keywords can be removed

\section{Introduction}
Explainability is an important concern in the deployment of robots in the real world, especially when these robots operate with and around humans \citep{sado2023explainable}. As robots are endowed with more complex decision-making, become more autonomous, and integrate further into people's daily lives, it becomes critical that those who interact with the robot or are affected by its decision-making understand how the robot behaves. In particular, a robot should be able to explain the causal reasons why it made a particular decision, or why something unexpected, such as a failure, occurred. In doing so, not only can users better understand the robot, but they can also more safely and efficiently integrate it into real-world applications \citep{miller2019explanation}. Indeed, making robots and other artificial intelligence (AI) systems more understandable through explainability is core to the creation of responsible and safe AI systems \citep{herrera2025responsible,walker2025harnessing}.\looseness=-1

In real-world decision-making tasks, many robots rely on structured control architectures. Behaviour trees (BTs) are a popular option for such an architecture, prevalent in robotics and across other fields such as artificial intelligence and computer gaming \citep{gugliermo2024evaluating}. In addition to their other qualities, such as reactivity and modularity, BTs are widely seen as an inherently interpretable, human-readable decision-making architecture \citep{iovino2022survey}, marking them as a promising tool in the design and implementation of transparent, interpretable robots in Human-Robot Interaction (HRI) scenarios.\looseness=-1

However, the inherent interpretability of BTs may not be sufficient for improving understandability, especially for non-expert users who may not be able to easily interpret representations of BTs. Thus, the problem of generating explanations from BTs has seen some recent attention \citep{han2021building,ogren2023creating,barkouki2024will,lemasurier2024reactive,tagliamonte2024generalizable}. Despite these efforts, the causal aspect of explainability has not been considered for BTs, and these approaches are unable to identify the reasons why a robot did or did not make a particular decision. A BT encodes a complex sequential relationship between BT nodes, and every state variable in the robot's environment may affect or be affected by a BT node. These causal relationships should be explicitly modelled to provide correct, causal explanations for BTs. Performing counterfactual reasoning (reasoning about hypothetical scenarios to contextualise real events) over such a model is considered an essential component of answering causal ```why'' questions~\citep{pearl2018book}. Additionally, domain knowledge, where available, can be used to expand these causal models. This knowledge can come in the form of models of the robot's environment, or prior knowledge about low-level decision-making components (such as BT nodes). To date, we are aware of no methods for generating causal explanations from BTs, except for generative (i.e. LLM-based) approaches, which are not guaranteed to correctly explain a particular decision~\citep{xu2024hallucination}. To ensure faithful understanding of the robot, provide safety assurances, and foster trust in the robot, it is essential that explanations are correct and respect the causal dynamics of the robot's decision-making.\looseness=-1

\begin{figure*}[t]
    \centering
    \includegraphics[width=\textwidth]{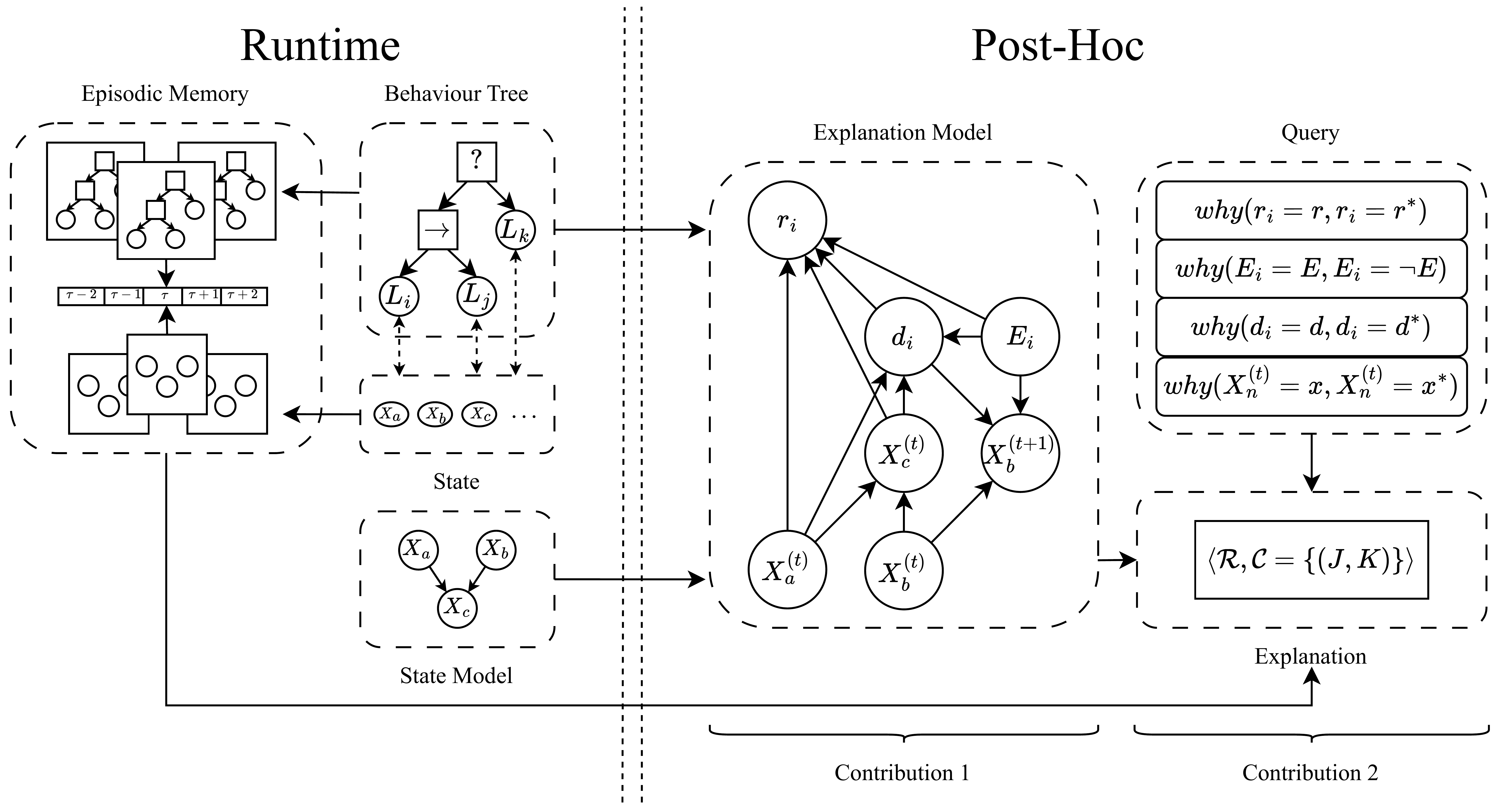}
    \caption{\footnotesize
    A diagram depicting the proposed architecture for explaining BT execution. At runtime, snapshots of the BT and the state are saved in an episodic memory (Section \ref{sec:architecture_execution}). Our contributions are twofold. The first contribution is an algorithm that builds a causal model (the explanation model) from the structure of the BT and state and decision-making knowledge (Section \ref{sec:architecture_causal_model}). Our second contribution is an algorithm to automatically generate explanations in real time by conducting a counterfactual search over the explanation model in order to satisfy a given query  (Section \ref{sec:architecture_explanation_generation}).}
    \label{fig:architecture_diagram}
\end{figure*}

Thus, the problem of providing correct, causal explanations of BTs presents a clear research gap. To address this gap, we propose a novel architecture (depicted in Figure \ref{fig:architecture_diagram}) that combines domain knowledge with the BT structure to create a causal model representation that enables the automatic generation of explanations in response to contrastive counterfactual queries. Our approach is capable of answering a wide range of queries regarding the success or failure of nodes in a BT, why a particular node was executed or not, the decisions the BT makes, and changes made to the state as a result of BT execution. Additionally, by leveraging an episodic memory that records past states and actions, our method can generate explanations based not only on the current state but also on previous states — both within a single BT tick and across longer temporal spans — supporting interactive follow-up questions. Our approach lays the foundation for behaviour tree-driven robots capable of interactively and automatically explaining the reasons for their decisions and behaviour, with the ultimate goal of producing transparent and understandable human-robot interactions.\looseness=-1

In summary, the contributions of this work are as follows:

\begin{enumerate}
    \item A method for combining structural and domain knowledge to create a causal model representation of the BT and the environment it interacts with (Algorithm \ref{alg:causal_graph}).
    \item An algorithm for conducting a counterfactual search over the generated causal model to produce a set of explanations in response to a contrastive query (Algorithm \ref{alg:counterfactual_search}).
\end{enumerate}

We evaluate our approach on a number of simulated behaviour trees (Section \ref{sec:evaluation}), including a simulated cognitive exercise domain, to demonstrate that the method is able to produce correct explanations in real time. A comparison with an LLM-based approach (Section \ref{sec:llm_baseline}) highlights the improvements our method offers in terms of explanation correctness and runtime. To ensure reproducibility, an implementation is provided at: \repourl.

\section{Background}\label{sec:background}
In this section, we provide some background on important concepts used in this work, namely behaviour trees (Section \ref{sec:background_behaviour_trees}) and causal models (Section \ref{sec:backbround_causal_models}).

\subsection{Behaviour Trees}\label{sec:background_behaviour_trees}
A BT is a directed tree consisting of internal control nodes and leaf execution nodes \citep{colledanchise2018behavior}. The tree functions by regularly signalling the root of the tree, which passes this signal (known as a \textit{tick}) to its children according to the type of node. Upon receiving a tick, a node is executed and returns either $\Success$ (indicating the node's function has been executed successfully), $\Failure$ (indicating the node's function has failed in some way), or $\Running$ (indicating the node's function is still executing). It can also be useful to regard nodes that have not been \textit{ticked} as returning an $\Invalid$ (or \textit{Idle}) status until such time that they are executed \citep{colledanchise2021implementation}.\looseness=-1

Leaf nodes can be categorised conceptually as either \textit{Action} nodes, which execute a command, or \textit{Condition} nodes, which check for a particular condition. Action nodes return $\Success$ if their command is executed successfully, $\Failure$ if the command fails, and $\Running$ if the command is still executing. Condition nodes, on the other hand, return $\Success$ if the condition is met and $\Failure$ otherwise, never returning $\Running$.

There are multiple commonly-used control nodes, of which we consider the two most fundamental - \textit{Sequence} and \textit{Fallback} nodes. \textit{Sequence} nodes, denoted using the \textrightarrow ~symbol, have two or more children which are executed in sequence, from left to right. The sequence node returns $\Success$ if all of its children return $\Success$, returning $\Failure$/$\Running$ if a single child returns $\Failure$/$\Running$.\looseness=-1

\textit{Fallback} nodes, also known as \textit{Selector} nodes, denoted using the $?$ symbol, have two or more children which are executed in sequence as with sequence nodes. However, the fallback node returns $\Failure$ only if all of its children return $\Failure$, otherwise returning $\Success$/$\Running$ if a single child returns $\Success$/$\Running$.

Following \cite{colledanchise2014behavior}, we can formally define a behaviour (sub)tree as the tuple $\mathcal{T}_i = \langle f_i, r_i, \Delta t \rangle$, where
\begin{itemize}
    \item $\Delta t$ is the time difference between ticking the tree and receiving an output.
    \item $f_i \colon \mathcal{S} \to \mathcal{S}$ is a function which determines the effect the activation of the subtree has on the state given an input state. In other words, $s^{(t_{k+1})} = f_i(s^{(t_k)})$, where $t^{(k+1)} = t^{(k)} + \Delta t$ and $k$ is a time step index. Here $\mathcal{S}$ is the set of all states and $s\in \mathcal{S}$ (see Section \ref{sec:definitions_decision_making}).
    \item $r_i \colon \mathcal{S} \to \{\Running,\Success,\Failure,\Invalid\}$ denotes the return function of the tree.
\end{itemize}

As nodes are organised in a tree structure, we can denote the set of children of $\mathcal{T}_i$ as $Ch(\mathcal{T}_i)$, and order all nodes in the tree such that $\mathcal{T}_i < \mathcal{T}_k$ indicates that $\mathcal{T}_i$ is to the left of $\mathcal{T}_j$. Similarly, $Pa(\mathcal{T}_i)$ denotes the parent node of $\mathcal{T}_i$. If two nodes share the same parent, they are referred to as siblings. As nodes are ordered from left to right, we call the child $\mathcal{T}_j \in Ch(\mathcal{T}_i)$ the ``left-most'' child of $\mathcal{T}_i$ if $\nexists \mathcal{T}_c \in Ch(\mathcal{T}_i)$ such that $\mathcal{T}_c < \mathcal{T}_j$.

\subsection{Causal Models}\label{sec:backbround_causal_models}
Causality is considered a central concept to explaining \textit{why} an event occurs, and modelling the causal relationships in a system with a \textit{causal model} is an important step in providing causal explanations \citep{pearl2018book}. Graphical causal models are particularly useful and represent causal relationships as edges in a directed acyclic graph (DAG). For the purposes of this work, we can define a causal model $M$ as a tuple $M = (\mathcal{G},\mathcal{R},\mathcal{F})$, where $\mathcal{G}=(\mathcal{V,\mathcal{E}})$ a DAG with nodes in $\mathcal{V}$ and edges in $\mathcal{E}$, $\mathcal{R}$ defines the range of each variable in $\mathcal{V}$, and $\mathcal{F}$ is a set of functions, one for each variable in $V \in \mathcal{V}$, mapping a product of all parents of $V$ in $\mathcal{G}$ to a value in $\mathcal{R}(V)$ \citep{miller2021contrastive}.

An important operator used in querying causal models is the \textit{Do} operator, which represents a causal intervention. The intervention $Do(X=x)$ asks the question ``What would happen if I force $X=x$ to be true regardless of the causal influences on $X$'' \citep{pearl2018book}. When applied to a model $\mathcal{M}$, $Do(X=x)$ deletes all incoming edges to $X$ and forces the value of the variable to $x$. The values of the causal descendants of $X$ can then be calculated using the existing values in the graph and the functions in $\mathcal{F}$.

%%%
%%% RELATED WORK
%%%

\section{Related Work}\label{sec:related_work}
The problem of automatically generating explanations for robots has seen some attention, adopting a wide variety of approaches from the field of explainable artificial intelligence. Some methods, such as that of \cite{gavriilidis2023surrogate}, learn interpretable surrogate models such as decision trees that approximate a robot's behaviour. Other methods leverage reinforcement learning, such as \cite{cruz2023explainable}, who explain a robot's learned policy in terms of task success probability, \cite{angelopoulos2025behind}, who incorporate user preferences in a reinforcement learning policy, and generate explanations when these preferences are violated. In this work, we focus specifically on counterfactual explanations, which identify important factors for a decision by perturbing an input in order to determine what changes would result in different decisions \citep{guidotti2022counterfactual}. Counterfactual reasoning has been argued to be an essential component of answering ``why'' questions whose answers leverage the causal reasons for a given decision or event \citep{miller2019explanation,pearl2018book}.\looseness=-1

Counterfactual explanations have seen some application in robotics domains. For example, \cite{gjaerum2023real} learn model trees as surrogates for a robot's decision-making and use them to generate counterfactual explanations. However, by not explicitly modelling causal assumptions on the state and decision-making, such a method may produce infeasible counterfactual scenarios that require additional feature engineering or optimisation constraints. Additionally, by not explicitly modelling causal relationships between features, an explanation might wrongly attribute a cause to a variable which is merely correlated to a true cause. Some methods counteract these problems by explicitly leveraging causal models when generating counterfactuals. For example, \cite{diehl2022did} learn a causal Bayesian network in simulation and then perform a breadth first search over the model to find valid counterfactuals, while \cite{love2024would} construct a simple causal graph modelling a human-robot interaction (HRI) scenario and generate sets of explanations by systematically intervening on nodes in the graph. However, these approaches do not consider the effects of prior states and decisions on the target failure or decision respectively, except where such history is encoded in the current state. In an extension to their previous work, \cite{diehl2023causal} account for temporal sequences of states and decisions by considering snapshots of states at discrete time intervals when learning the structure of the causal Bayesian network. Nevertheless, this method remains focused on explaining failures rather than the reasons for decisions and states. It also requires learning a causal model in simulation, without exploiting the structure of the robot's control architecture to automatically define causal relationships in the robot's behaviour.

As our work specifically leverages the structure of behaviour trees in order to construct a causal model and generate counterfactual explanations,  we now discuss some existing approaches linking explainability and BTs. BTs are considered to be inherently more interpretable than black box models such as neural networks, and thus several approaches focus on learning BTs as surrogate models \citep{potteiger2023safe,wan2024unraveling}. While this claim is certainly true, and much can be understood by examining depictions of a BT state, it is nevertheless useful to explicitly generate explanations from BTs, especially for non-expert users. There are some works in this area, such as the foundational work of \cite{han2021building}, who introduce several algorithms that are capable of answering a variety of questions about a behaviour tree at runtime, such as ``What are you doing?'', answered by describing the current node in execution, ``Why are you doing this?'', answered by describing ancestors of the current node in the tree, and ``What is your (sub)goal?'', answered by describing the entire (sub)tree. Of these questions, only ``Why are you doing this?'' is causal in nature, and then only querying the execution of the currently executing BT node, answering by referring to the execution of a single ancestor node. This approach is not designed to cater to a wider variety of causal queries targeting decisions, node return statuses or state variable values. It is also not designed to give causal explanations for events that did not occur (e.g. answering ``Why didn't you do that?'').\looseness=-1

The approach by \cite{han2021building} is expanded on by \cite{barkouki2024will}, who generate explanations by predicting future actions using the structure of a BT, answering questions such as ``What will you do next if the current action succeeds?''. \cite{ogren2023creating} extend the work to backward chained BTs, producing explanations that reference both successful and unsuccessful goals as well as a trace of node executions up the tree from the currently executing BT node. By constructing a BT as a ``Make-sure BT'', condition nodes which check action preconditions allow for state variables references by preconditions to be included in the explanation. \cite{lemasurier2024reactive} also expand on the work of \cite{han2021building} by introducing assumption checker nodes to the BT, which track whether or not key assumptions have been violated. If such assumptions have been violated, the robot can predict upcoming failures and proactively explain them. While these approaches are indeed useful, they do not address the causes of decisions and behaviour beyond the sequence of subtree executions (and, in the case of \cite{ogren2023creating}, certain state variables representing preconditions). For an illustrative comparison between our method and those of \cite{han2021building} and \cite{ogren2023creating}, see Section \ref{sec:comparison_with_literature}.\looseness=-1

Recently, LLMs have been presented as a tool for both generating and formatting explanations of robotics systems. For example, \cite{frering2025integrating} allow users to interface with an LLM which can produce explanations for the behaviour of a belief-desire-intention (BDI) agent in response to natural language queries. The explanation prompt includes information on the BDI agent's ``mind state'' and the interaction history of the system with the user. \cite{gebelli2025personalised} provide an LLM with both static (e.g. behaviour rules) and dynamic knowledge (e.g. execution logs), as well as a user profile, in order to generate personalised natural language explanations.
Of particular relevance to our work is that of \cite{tagliamonte2024generalizable}, who provide a general framework for answering user queries about a BT using an LLM, providing information such as the BT (encoded in YAML format), its current state, and any violated assumptions in the form of a prompt. Such an approach is clearly attractive, as both queries and explanations can be provided in natural language. However, there are no guarantees that the explanation provided by an LLM corresponds to the reasoning of the BT. Furthermore, results in a follow-up user study found the generated explanations to be less understandable and trustworthy compared to a templated approach \citep{lemasurier2024templated}. In contrast, our approach explicitly incorporates the structure of the BT and domain knowledge about tree nodes and the state in order to provide explanations that are faithful to the BT's reasoning and the dynamics of the environment, as demonstrated in Section \ref{sec:evaluation}.\looseness=-1

Of all the methods presented in this section, only the LLM-based approaches are capable of responding to the same types of contrastive queries that our method targets, due to their ability to process arbitrary textual inputs to produce coherent natural language outputs. We therefore directly compare our method to an end-to-end LLM-based approach, representative of this family of approaches, in Section \ref{sec:llm_baseline}.

%%%
%%% DEFINITIONS
%%%

\section{Definitions}\label{sec:definitions}
In this section, we present the formalism we use to specify counterfactual queries and explanations (Section \ref{sec:definitions_counterfactuals}), as well as our extension of the formal definition of a behaviour tree (Section \ref{sec:extending_bts}).\looseness=-1

\subsection{Decision-Making}\label{sec:definitions_decision_making}
Decision-making problems are often expressed in terms of states, configurations of the environment, actions, and decisions taken by the system \citep{russel2009artificial}. In this work, we assume that the set of all states, termed the \textit{state space} and denoted $\mathcal{S}$, can be factored into a configuration of state variables $\mathcal{S} = \prod_{n=1}^{|\mathcal{X}|}X_n$, where $\mathcal{X}$ is the set of state variables and $X_n \in \mathcal{X}$. Thus, each state can be expressed as $s = \langle x_1,...,x_n\rangle$, where each $x_n$ denotes a particular value of $X_n$. Similarly, we term the set of all actions the \textit{action space}, denoted $\mathcal{A}$. We assume that the null action $\varnothing$ - that is, the system doing nothing - is a member of $\mathcal{A}$.

\subsection{Counterfactual Queries and Explanations}\label{sec:definitions_counterfactuals}
In this work, we seek to answer questions of the form ``Why $A$ and not $B$?''. Following \cite{miller2021contrastive}, we assume that the true event $A$ (the \textit{fact}) and untrue event $B$ (the \textit{foil}) refer to the same variable or set of variables, and that they are mutually exclusive (i.e. they cannot both be true). We denote the query using the notation $\Why(A,B)$.\looseness=-1

To answer such a query, we define a counterfactual explanation as a tuple $\langle \mathbf{R},\mathbf{C}\rangle$. $\mathbf{R}$ denotes the \textit{reason set}, which is a set of true statements or \textit{reasons} that each explain the fact $A$. $\mathbf{C}$ denotes the \textit{counterfactual set}, which is a set of counterfactual tuples $(\mathbf{J},\mathbf{K})$, where $\mathbf{J}$ is a statement that, if true, implies $\mathbf{K}$ is true, where $\mathbf{K} \implies B$. Each $(\mathbf{J},\mathbf{K}) \in \mathbf{C}$ must correspond to a reason in $\mathbf{R}$ (and inversely, each reason should have at least one corresponding counterfactual), such that $\mathbf{J}$ is mutually exclusive with the reason.

As an example, consider a robot running a cognitive exercise, in which a user is shown a sequence of symbols by the robot and must repeat the sequence from memory (this use case, known as the \textit{serial recall} task, is properly introduced in Section \ref{sec:evaluation_serial_recall}). Suppose that, after receiving an incorrect sequence from the user, the robot decides to end the task. A caregiver reviewing the interaction might ask ``Why did the robot end the task rather than repeat the sequence for the user?''. As a formal query, this can be posed as $\Why(fact: d = EndTask, foil: d = RepeatSequence)$. Suppose that the robot provides the formal explanation $\langle \mathbf{R}: UserFrustration = 0.8, \mathbf{C}: (\mathbf{J}: UserFrustration \leq 0.7, \mathbf{K}: d = RepeatSequence) \rangle$. This explanation would, translated to natural language, read as: ``The robot ended the current task because the user's frustration level was $0.8$. If the user's frustration level was $0.7$ or lower, the robot would have repeated the sequence for the user.''\looseness=-1

\subsection{Extending Behaviour Trees}\label{sec:extending_bts}
While the formal representation of BTs presented in Section \ref{sec:background_behaviour_trees} encodes the sequence of decisions made by the system, it does not in itself capture the relationships between the concepts involved in the decisions, such as which parts of the state are directly related to particular parts of the decision-making architecture. Thus, to obtain a rich set of counterfactual explanations from a BT which also incorporates this domain knowledge about the relationship between the state and the decision-making (e.g. the input/output of each node), we extend the definition of \cite{colledanchise2014behavior} within the framework of the decision-making problem outlined in Section \ref{sec:definitions_decision_making}. In particular, we define a leaf node $L_i$, an extension of the generic subtree $\mathcal{T}_i$, as a tuple $\langle \mathcal{X}_i, \mathcal{Y}_i, \mathcal{A}_i, f_i, p_i, d_i, r_i, \Delta t \rangle$, whose elements are defined as follows:\looseness=-1

\begin{itemize}
    \item $\mathcal{X}_i \subseteq \mathcal{X}$ denotes the set of state variables taken as input by the leaf node (the input variables). In practice, this set could represent blackboard variables read by the node. From $\mathcal{X}_i$, we can define an input state space $\mathcal{I}_i = \prod_{n=1}^{|\mathcal{X}_i|}X_n$.
    \item $\mathcal{Y}_i \subseteq \mathcal{X}$ denotes the set of state variables whose values can be directly changed by the leaf node (the output variables). In practice, this set could represent the blackboard variables written to by the node. From $\mathcal{Y}_i$, we can define an output state space $\mathcal{O}_i = \prod_{n=1}^{|\mathcal{Y}_i|}Y_n$.
    \item $\mathcal{A}_i \subseteq \mathcal{A}$ denotes the set of actions that can be selected by the node. We assert that the null action $\varnothing \in \mathcal{A}_i$.
    \item $p_i: \mathcal{I}_i \to \mathcal{O}_i$ represents the direct effect of the nodes' execution on the output variables in $\mathcal{Y}_i$, given the input variables in $\mathcal{X}_i$. 
    \item $d_i: \mathcal{I}_i \to \mathcal{A}_i$ denotes the decision-making function (i.e. policy) of the node, determining which action is selected when the node is executed. If the node is not executed, or if the node cannot select an action (such as with Condition nodes), then $d_i = \varnothing$.
    \item $f_i$, $r_i$ and $\Delta t$ retain their definitions from Section \ref{sec:background_behaviour_trees}.
\end{itemize}

For convenience, we can also define the statement $E_i$, which is true if $L_i$ is executed and false otherwise.

We motivate our decision to include $\mathcal{A}_i$ and $d_i$ in this extended definition by noting that querying the behaviour and actions taken by a robot is an important aspect of explainable robotics \citep{wachowiak2024people}. If a node or subtree can only ever execute one action primitive, then this addition is redundant. However, if a node is capable of executing more than one action, our formulation is required in order to explain why one action was executed instead of another (see Section \ref{sec:architecture_explanation_generation}). This distinction is also useful for differentiating between intentions (decisions) and executions, where the actual execution of a decision may fail.

Here we also note an important assumption we make going forward: that the functions $p_i$, $d_i$, and $r_i$ are deterministic. This assumption is necessary for the counterfactual search we present in Section \ref{sec:architecture_explanation_generation} (Algorithm \ref{alg:counterfactual_search}), as it repeatedly executes $p_i$, $d_i$, and $r_i$, which must return consistent results. Similarly, $f_i$ should be deterministic in order to ensure that the explanations are consistent with the real environment. For a discussion on this limitation and on future work arising from it, see Section \ref{sec:limitations}.

%%%
%%% ARCHITECTURE
%%%

\section{Architecture}\label{sec:architecture}

\begin{figure}[t]
    \centering
    \includegraphics[width=\textwidth]{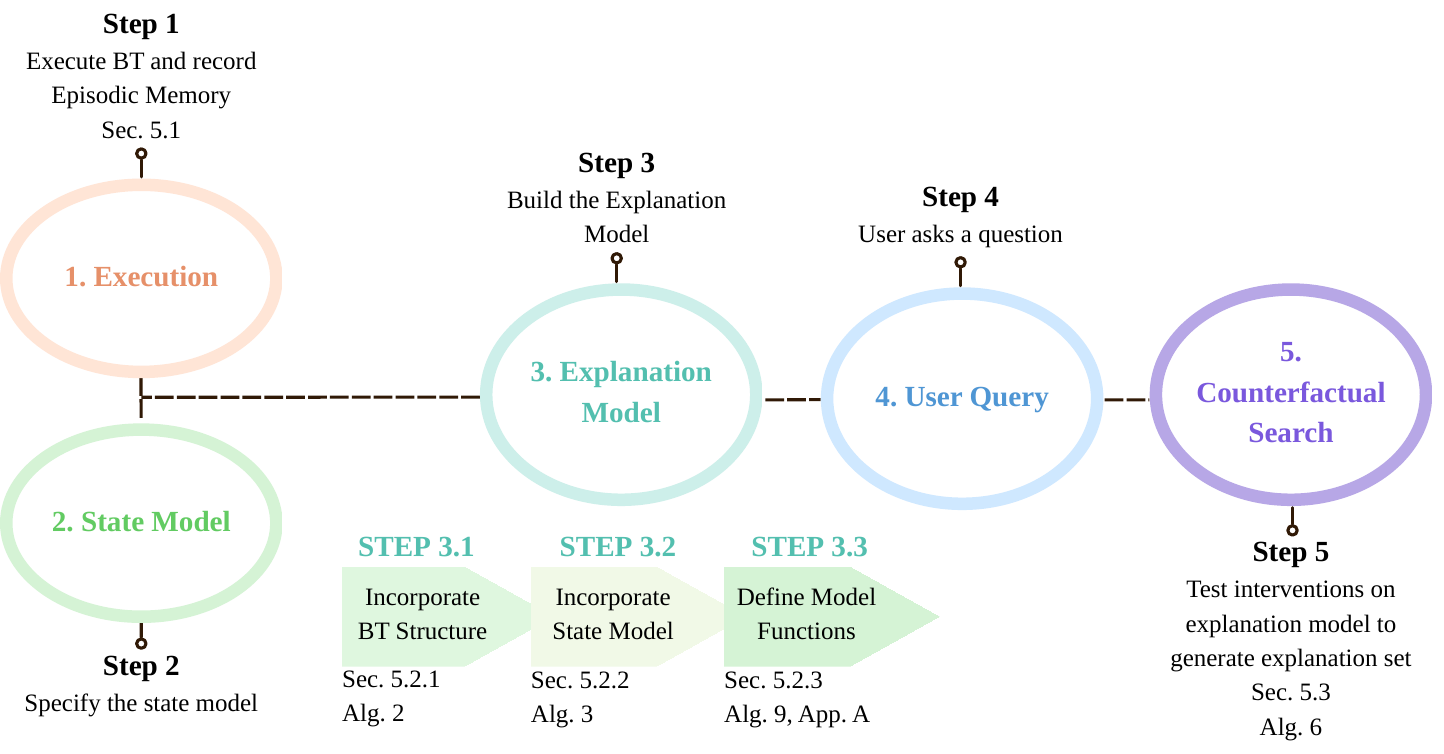}
    \caption{\footnotesize
    A high-level overview of the implementation of our approach. The first two steps, which may occur in any order, are to 1) execute the BT and record the episodic memory in a suitable format, and 2) specify the domain knowledge to be injected into the system in the form of a state model $\mathcal{M}_\mathcal{S}$. Subsequently, our framework 3) builds an explanation model, which is a causal model representing the BT and the state which can be used to generate explanations. Constructing the model contains three substeps, namely 3.1) incorporating the BT structure into the model structure, 3.2) incorporating the state model into the model structure, and 3.3) defining variable ranges and functions for the model. Finally, whenever 4) a user asks a contrastive question (Section \ref{sec:definitions_counterfactuals}), our framework 5) conducts a counterfactual search over the explanation model to generate a set of explanations that answer the user's query.}
    \label{fig:high_level_flowchart}
\end{figure}

Having provided the necessary definitions, we now present our methodology to generate explanations for systems (such as robots) whose decision-making is governed by a BT. Our architecture, depicted in Figure \ref{fig:architecture_diagram}, relies on episodic memory stored at runtime (Section \ref{sec:architecture_execution}). It operates by constructing a graphical model (explanation model) that during or post-execution captures the causal relationships derived from both the BT structure and domain knowledge about the state and decision-making (Section \ref{sec:architecture_causal_model}). Using both the explanation model and the episodic memory, our method then performs a counterfactual search to automatically generate explanations in real time that satisfy a given query (Section \ref{sec:architecture_explanation_generation}). A high-level flowchart depicting this process is depicted in Figure \ref{fig:high_level_flowchart}. In addition to presenting the algorithms that make up the core contributions of this work, we also discuss some properties thereof and conduct complexity analyses in Section \ref{sec:architecture_properties}.\looseness=-1

\subsection{Execution}\label{sec:architecture_execution}
\textit{Episodic memory} - a detailed memory of states, actions and other relevant information over time -  is considered an essential component to explaining why a particular event, such as a robot decision or BT state, occurs \citep{langley2017explainable,dechant2023risks}. 

Our method relies on the system maintaining such an episodic memory which captures snapshots of the BT and the values of state variables, in order to be able to reconstruct the context of a queried event and be able to consider counterfactual scenarios deriving from those circumstances. In particular, we record all changes to the state variables in $\mathcal{X}$, as well as the return status (and action, if applicable) of a node when it is executed. In this way, the discrete unit of time is defined by the execution of exactly one node.\looseness=-1

All BTs in this work are implemented in the \textit{py\_trees} library \citep{pytrees2025}, with memories recorded by a custom visitor.\looseness=-1

\subsection{Building the Causal Model}\label{sec:architecture_causal_model}

In order to generate counterfactual explanations, our method requires a graphical model that encodes the causal relationships present between state variables in $\mathcal{X}$ and within the BT (between the execution $E_i$, return status $r_i$ and decision $d_i$ for each BT node $\mathcal{T}_i$). Therefore, one of the core contributions of this work is an algorithm to automatically construct such a causal model, which we term the \textit{explanation model}. 

Recall from Section \ref{sec:backbround_causal_models} that such a model is specified by the tuple $M = (\mathcal{G},\mathcal{R},\mathcal{F})$. The graphical structure $\mathcal{G} = (\mathcal{V},\mathcal{E})$ is derived from both the known structure of the BT (Section \ref{sec:architecture_causal_model_bt_structure}) as well as domain knowledge about the state and decision-making (Section \ref{sec:architecture_causal_model_domain_knowledge}), as represented in Algorithm \ref{alg:causal_graph} (a visual representation is also provided in Figure \ref{fig:flowchart_build_model}, Appendix \ref{app:algorithms}). Given this structure, we can then determine the ranges $\mathcal{R}$ and functions $\mathcal{F}$ for each node in the model (Section \ref{sec:architecture_causal_model_functions}).

\begin{algorithm}
    \caption{Building the explanation graph}\label{alg:causal_graph}
    \begin{algorithmic}[1]
    \Function{BuildExplanationGraph}{}
        \State $\mathcal{G} \gets $ \Call{GraphFromStructure}{$ $} \Comment{Alg. \ref{alg:cm_from_bt}}
        \State $\mathcal{G} \gets $ \Call{AddDomainKnowledge}{$\mathcal{G}$} \Comment{Alg. \ref{alg:cm_from_dk}}
        \State \textbf{return} $\mathcal{G}$
    \EndFunction
    \end{algorithmic}
\end{algorithm}

As a BT is a sequential decision-making structure whose nodes are executed at different points in time, the explanation model must be capable of representing a temporal sequence of events rather than a single moment in time. In our construction, the explanation model represents the system for the duration of a single, arbitrary ``tick'' of the BT (i.e. from the execution of the tree root to the moment it returns a valid return status).

\subsubsection{Incorporating Behaviour Tree Structure}\label{sec:architecture_causal_model_bt_structure}
As mentioned before, some causal relationships in $\mathcal{G}$ can be determined purely from the structure of the BT. Variables can be inserted into the explanation model for each node in the tree. We devise different methods for relating these variables in the explanation graph, depending on the type of node and its relationship with its parent and siblings. The algorithm for determining the partial causal graph from the BT structure is presented in Algorithm \ref{alg:cm_from_bt} (a full version is given in Algorithm \ref{alg:cm_from_bt_full}, Appendix \ref{app:algorithms}).\looseness=-1

\begin{algorithm}
    \caption{Constructing the explanation graph from the BT structure (full version in Algorithm \ref{alg:cm_from_bt_full}, Appendix \ref{app:algorithms})}\label{alg:cm_from_bt}
    \begin{algorithmic}[1]
    \Function{GraphFromStructure}{ } \funclabel{algf:graph_from_structure} \label{algl:graph_from_structure}
        \State $\mathcal{G} \gets \varnothing$
        \State $\mathcal{G} \gets$ \Call{AddVariablesFromBTNodes}{$\mathcal{G}$} \Comment{Insert graph nodes for each BT node}\label{algl:add_variables_from_bt_nodes_line}
        \For {node $\mathcal{T}_i$ in the BT} 
            \State Add edge $E_i \to r_i$ to $\mathcal{E}$\label{algl:add_E_to_R_edge}
            \If{$\mathcal{T}_i$ is a leaf node}
                \State $\mathcal{G} \gets$ \Call{AddLeafEdges}{$\mathcal{G}$, $\mathcal{T}_i$} \Comment{Depicted in Figure \ref{fig:structure_leaf_subgraph}}\label{algl:add_leaf_edges_line}
            \Else
                \State $\mathcal{G} \gets$ \Call{AddCompositeEdges}{$\mathcal{G}$, $\mathcal{T}_i$} \Comment{Depicted in Figure \ref{fig:structure_composite_subgraph}}\label{algl:add_composite_edges_line}
            \EndIf
            \If{$\mathcal{T}_i$ is not the tree root}
                \State $\mathcal{G} \gets$ \Call{AddParentSiblingEdges}{$\mathcal{G}$, $\mathcal{T}_i$} \Comment{Depicted in Figures \ref{fig:structure_left_child} and \ref{fig:structure_other_child}}\label{algl:add_parent_sibling_edges_line}
            \EndIf
        \EndFor
        \State \textbf{return} $\mathcal{G}$
    \EndFunction
    \end{algorithmic}
\end{algorithm}

\begin{figure}
    \centering
    \begin{subfigure}{.28\textwidth}
      \centering
      \includegraphics[width=\textwidth]{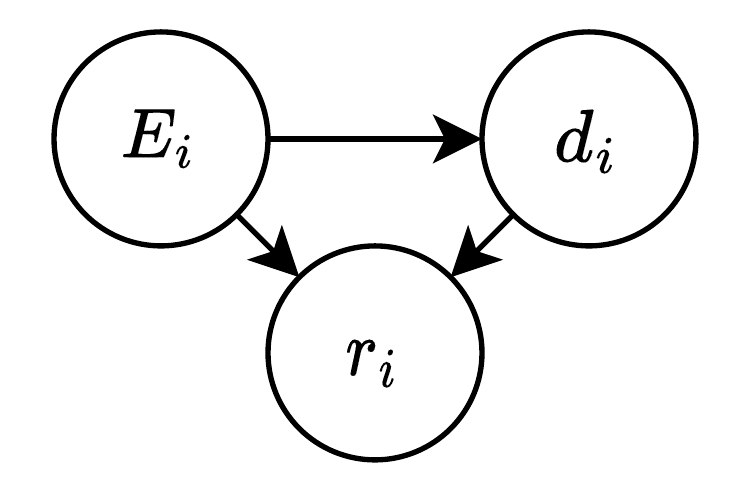}
      \caption{$\mathcal{T}_i$ is a leaf node.}
      \label{fig:structure_leaf_subgraph}
    \end{subfigure}%
    \hfill
    \begin{subfigure}{.40\textwidth}
      \centering
      \includegraphics[width=\linewidth]{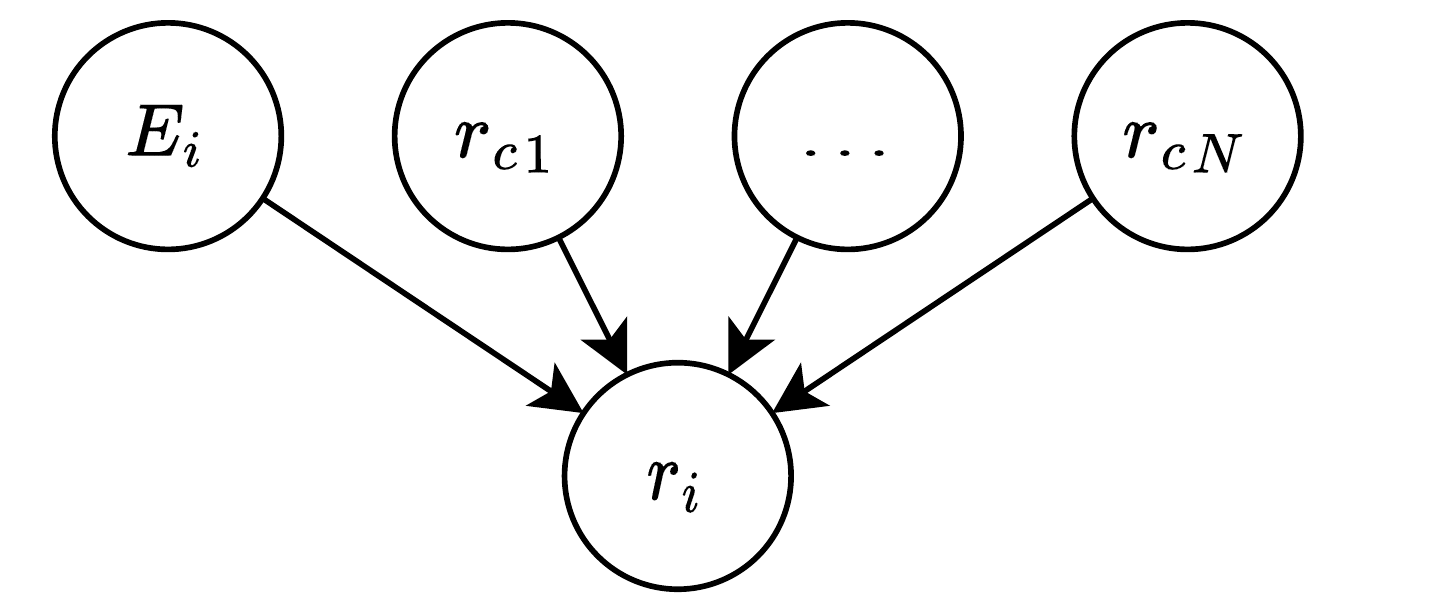}
      \caption{$\mathcal{T}_i$ is a composite node.}
      \label{fig:structure_composite_subgraph}
    \end{subfigure}%
    \hfill
    \begin{subfigure}{.15\textwidth}
      \centering
      \includegraphics[width=\linewidth]{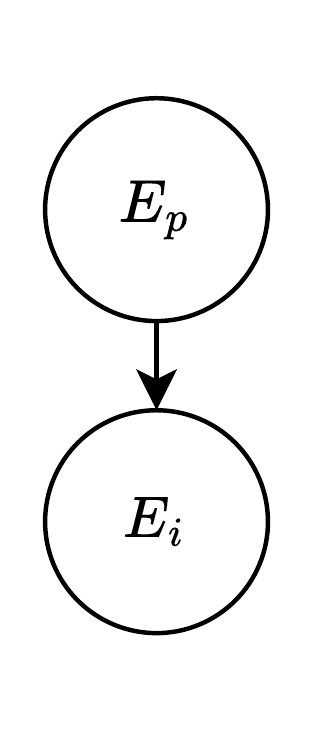}
      \caption{$\mathcal{T}_i$ is the left-most child of $\mathcal{T}_p$.}
      \label{fig:structure_left_child}
    \end{subfigure}%
    \hfill
    \begin{subfigure}{.15\textwidth}
      \centering
      \includegraphics[width=\linewidth]{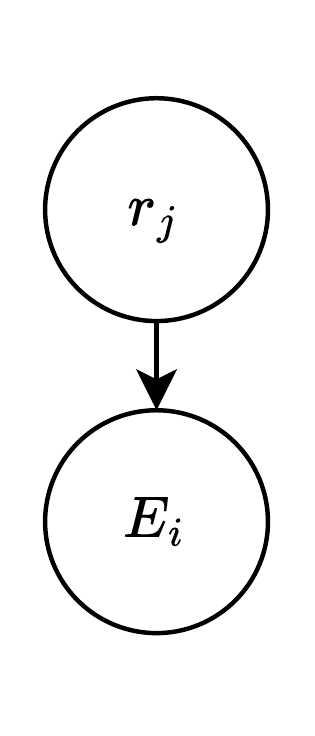}
      \caption{$\mathcal{T}_i$ is not the left-most child of $\mathcal{T}_p$.}
      \label{fig:structure_other_child}
    \end{subfigure}%
    \caption{Explanation subgraphs of $\mathcal{T}_i$. For leaf nodes, $E_i$, $d_i$ and $r_i$ are related as in (a). For composite nodes, $r_i$ is related to $E_i$ and ${r_c}_1$,...,${r_c}_N$ as in (b), where ${\mathcal{T}_c}_1,...,{\mathcal{T}_c}_N \in Ch(\mathcal{T}_i)$. When $\mathcal{T}_i$ is the left-most child of its parent $\mathcal{T}_p = Pa(\mathcal{T}_i)$, $E_i$ is dependent on $E_p$, as in (c). Otherwise, $E_i$ is dependent on $r_j$, as in (d), where $\mathcal{T}_j$ is the sibling of $\mathcal{T}_i$ immediately to its left.}
    \label{fig:structure_add_nodes}
\end{figure}

First of all, we note that for any node $\mathcal{T}_i$, it can only return a status in $\{\Running,\Success,\Failure\}$ if it is executed, otherwise returning $\Invalid$. Thus, $r_i$ is dependent on $E_i$ (Algorithm \ref{alg:cm_from_bt}, line \ref{algl:add_E_to_R_edge}). For a leaf node $L_i$ for which $d_i$ is defined (i.e. an action node), we note some additional relationships (Algorithm \ref{alg:cm_from_bt}, line \ref{algl:add_leaf_edges_line}, depicted in Figure \ref{fig:structure_leaf_subgraph}). The decision $d_i$ is dependent on $E_i$, as $d_i = \varnothing$ by definition if the node is not executed. If the node's decision-making is implemented in two stages, as we do in this work, such that a decision is first made and then acted out in the environment, then we can say that $r_i$ depends on $d_i$. For implementations where this does not hold true, then this relationship may be ignored.

For a composite node $\mathcal{T}_i$, we note that $r_i$ is dependent on the return statuses of its children ${\mathcal{T}_c}_1,...,{\mathcal{T}_c}_N \in Ch(\mathcal{T}_i)$ (Algorithm \ref{alg:cm_from_bt}, line \ref{algl:add_composite_edges_line}, depicted in Figure \ref{fig:structure_composite_subgraph}), following the rules outlined in Section \ref{sec:background_behaviour_trees}.\looseness=-1

Finally, we consider the influences on whether or not a node $\mathcal{T}_i$ is executed ($E_i$). If $\mathcal{T}_i$ is the root of the tree, then we assume $E_i$ to be true. Otherwise, the causal relationships depend on the position of $\mathcal{T}_i$ among its siblings (Algorithm \ref{alg:cm_from_bt}, line \ref{algl:add_parent_sibling_edges_line}, depicted in Figures \ref{fig:structure_left_child} and \ref{fig:structure_other_child}). Suppose $\mathcal{T}_p = Pa(\mathcal{T}_i)$. If $\mathcal{T}_i$ is the left-most child of $\mathcal{T}_p$, then for both sequence and fallback nodes, $\mathcal{T}_i$ is executed first and thus $E_i = E_p$ (Figure \ref{fig:structure_left_child}). Otherwise, there must exist some sibling $\mathcal{T}_j$ immediately to the left of $\mathcal{T}_i$. We observe then that $E_i$ is dependent on $r_j$, with the function governing that relationship depending on whether $\mathcal{T}_p$ is a sequence or fallback node (see Section \ref{sec:architecture_causal_model_functions}), as these nodes execute their children from left to right with stopping conditions dependent on child return statuses (Figure \ref{fig:structure_other_child}).

\subsubsection{Incorporating Domain Knowledge}\label{sec:architecture_causal_model_domain_knowledge}

\begin{figure}
    \centering
    \begin{subfigure}{.20\textwidth}
      \centering
      \includegraphics[width=\textwidth]{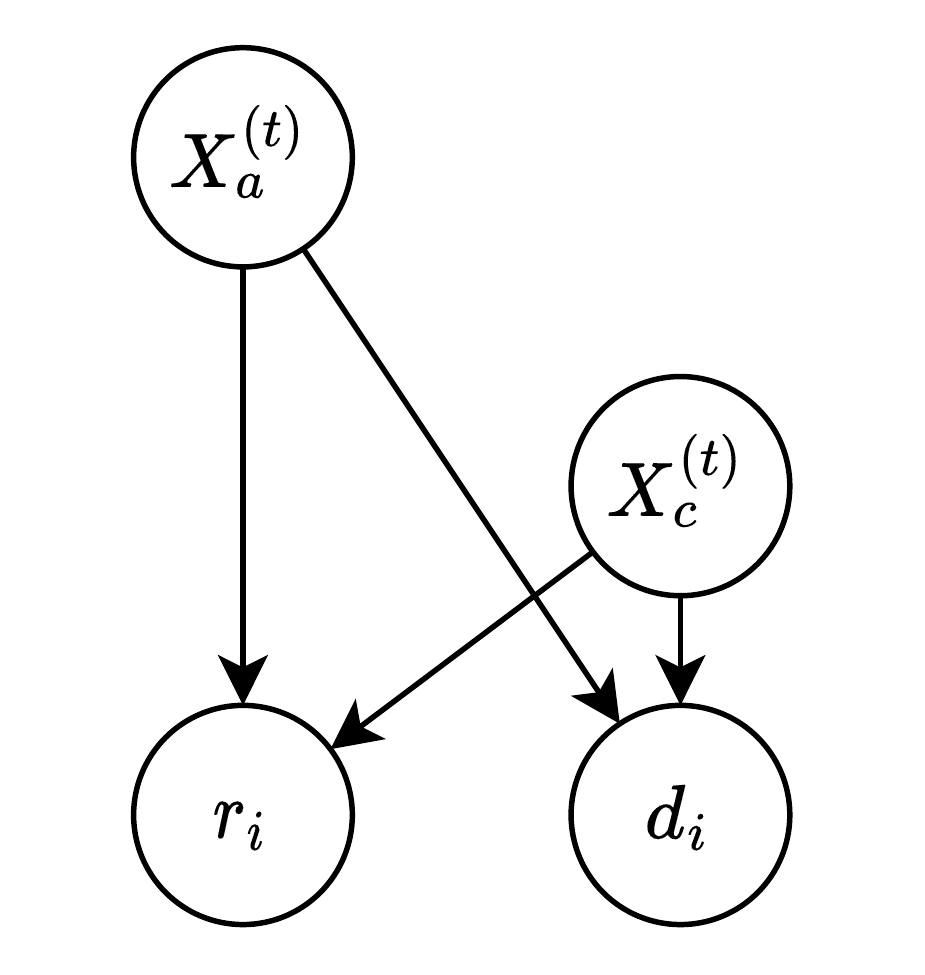}
      \caption{Input variables in $\mathcal{X}_i$ are added to the graph and linked to $r_i$ (and $d_i$ for action nodes).}
      \label{fig:graph_from_dk_step_1}
    \end{subfigure}%
    \hfill
    \begin{subfigure}{.24\textwidth}
      \centering
      \includegraphics[width=\linewidth]{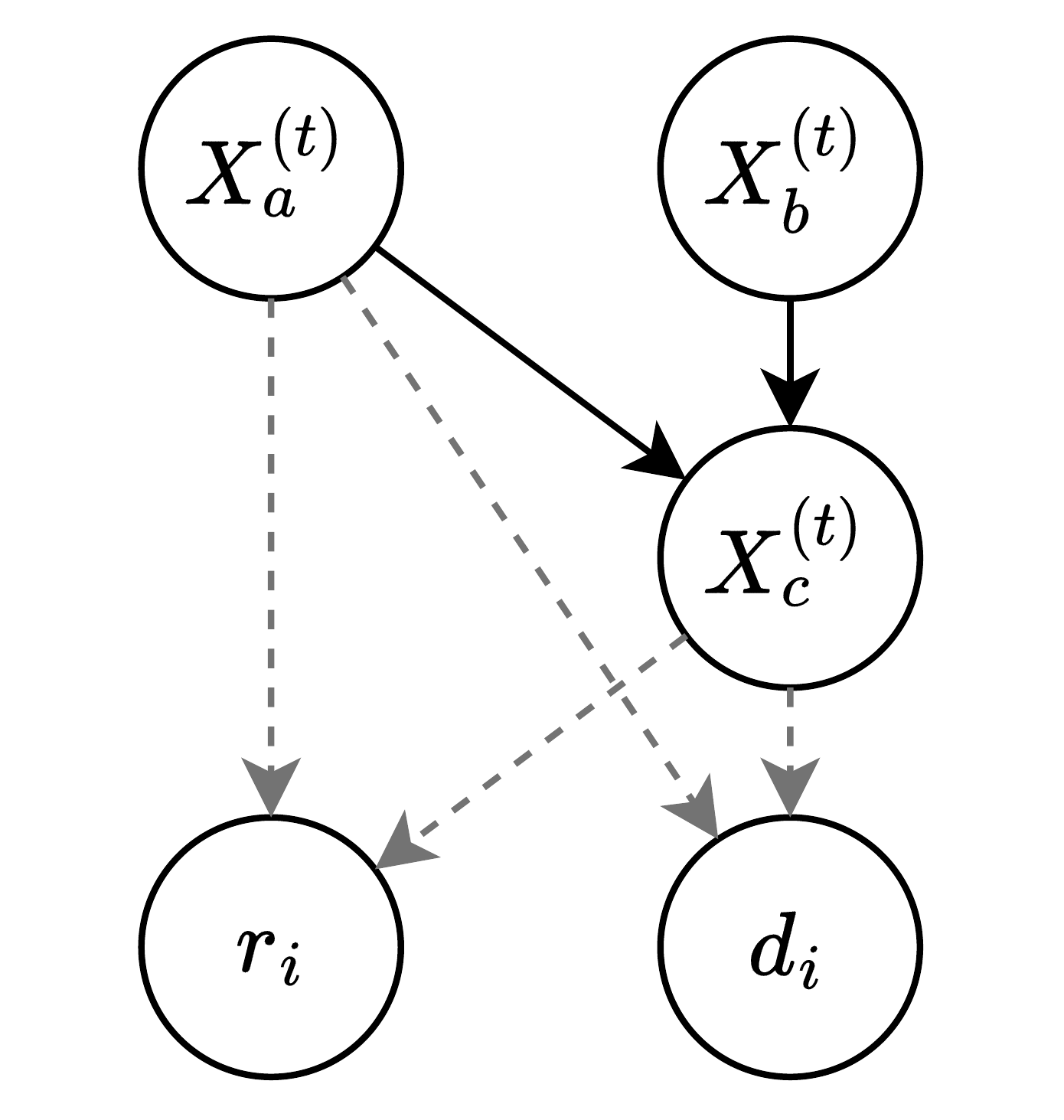}
      \caption{The causal ancestors defined by $\mathcal{G}_\mathcal{S}$ are added to the graph and linked to their descendants.}
      \label{fig:graph_from_dk_step_2}
    \end{subfigure}%
    \hfill
    \begin{subfigure}{.26\textwidth}
      \centering
      \includegraphics[width=\linewidth]{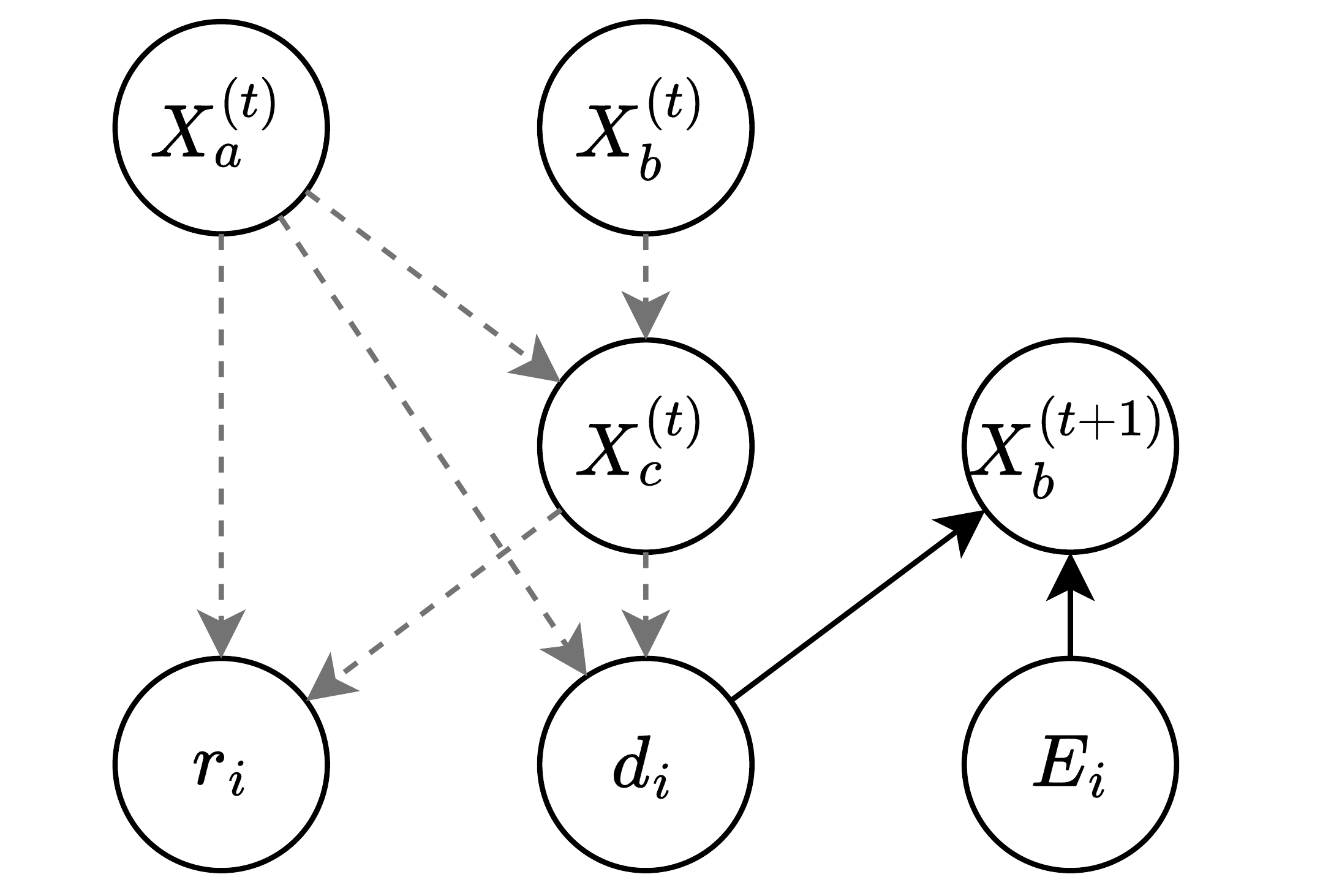}
      \caption{New temporal versions of output variables in $\mathcal{Y}_i$ are added to the graph and linked to $d_i$ and $E_i$.}
      \label{fig:graph_from_dk_step_3}
    \end{subfigure}%
    \hfill
    \begin{subfigure}{.26\textwidth}
      \centering
      \includegraphics[width=\linewidth]{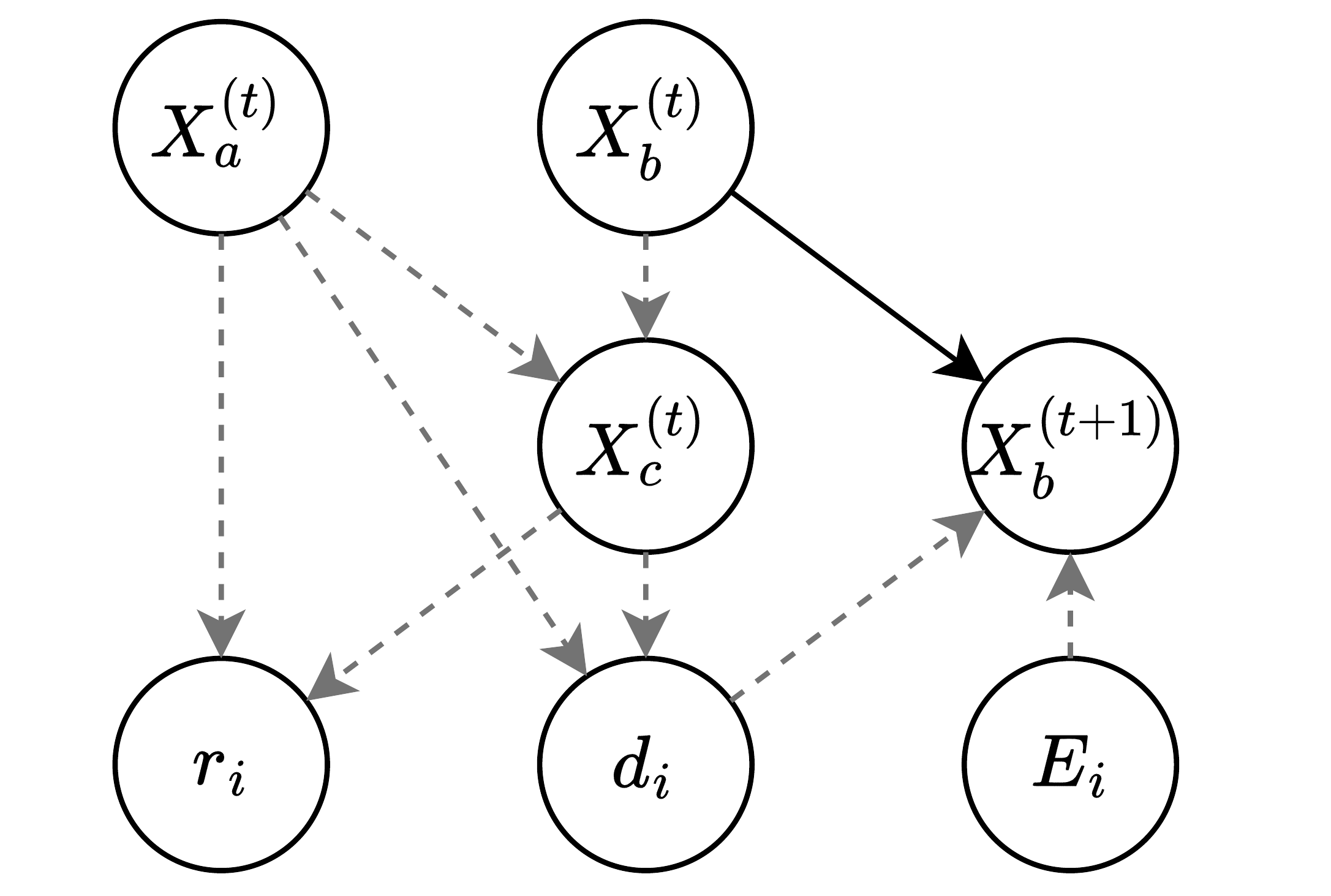}
      \caption{New temporal versions of top-level variables are linked to the old versions.}
      \label{fig:graph_from_dk_step_4}
    \end{subfigure}%
    \caption{An example of the edges added by Algorithm \ref{alg:cm_from_dk} for one $L_i$. In this example, the state graph $\mathcal{G}_{\mathcal{S}}$ is identical to the one in Figure \ref{fig:state_graph_example}, $\mathcal{X}_i=\{X_a,X_c\}$, and $\mathcal{Y}_i=\{X_b\}.$}
    \label{fig:graph_from_dk}
\end{figure}

Once the partial explanation graph has been constructed from the known BT structure, further causal relationships can be inserted from domain knowledge. This process is depicted in Algorithm \ref{alg:cm_from_dk}, and an example is provided in Figure \ref{fig:graph_from_dk}. We distinguish between two types of domain knowledge: \textit{state knowledge}, regarding the causal relationships between variables in $\mathcal{X}$, and \textit{decision-making knowledge}, regarding elements of the decision-making process on the level of individual nodes in the BT.

\begin{figure}[t]
    \centering
    \includegraphics[width=0.4\textwidth]{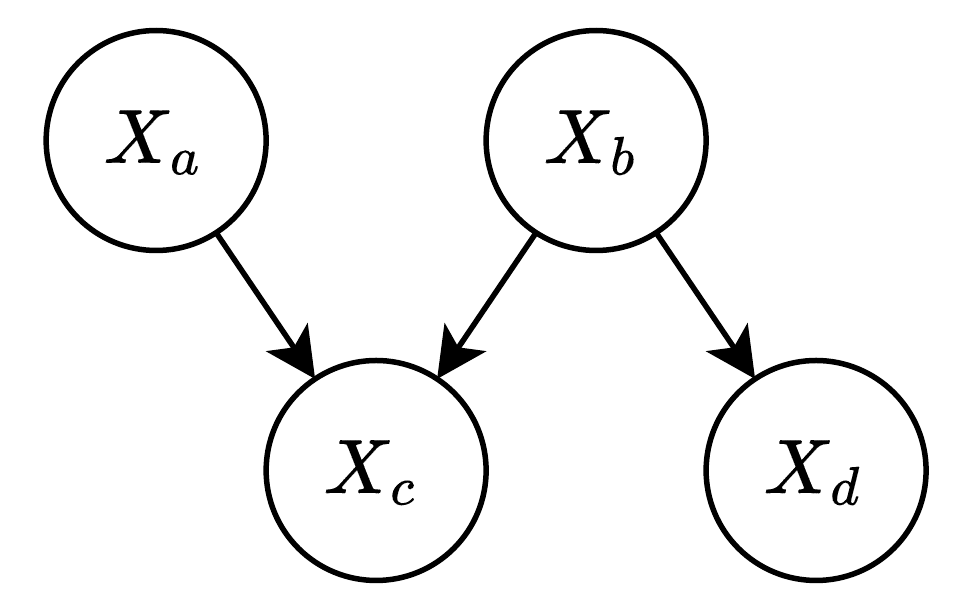}
    \caption{\footnotesize
    An example state graph with four variables in $\mathcal{X}$.}
    \label{fig:state_graph_example}
\end{figure}

State knowledge is represented by a state model $M_{\mathcal{S}} = (\mathcal{G}_{\mathcal{S}},\mathcal{R}_\mathcal{S},\mathcal{F}_{\mathcal{S}})$. Inserting this knowledge into the explanation model allows for the generated explanations to include a wider range of causes (e.g. including not just an input variable but its causal ancestors as well) and to account for the delayed temporal effects of BT node executions and decisions (e.g. a decision impacting a state variable whose causal descendant subsequently impacts another event). An example state graph is provided in Figure \ref{fig:state_graph_example}.\looseness=-1

Decision-making knowledge is represented by the extended BT sets and functions defined in Section \ref{sec:extending_bts}. In particular, $\mathcal{X}_i$, $\mathcal{Y}_i$ and $\mathcal{A}_i$ must be provided. The functions $p_i$, $d_i$ and $r_i$ are not necessarily known, but can be evaluated for any given input by executing the BT node.\looseness=-1

\begin{algorithm}
    \caption{Completing the explanation graph with domain knowledge}\label{alg:cm_from_dk}
    \begin{algorithmic}[1]
    \Function{AddDomainKnowledge}{$\mathcal{G}$} \funclabel{algf:add_domain_knowledge} \label{algl:add_domain_knowledge}
        \State Define mapping $\tau:\mathcal{X} \to \mathbb{Z}$
        \State $\forall X \in \mathcal{X}$, add $X^{(0)}$ to $\mathcal{V}$ and set $\tau(X) \gets 0$ \label{algl:add_inital_variables_line}
        \State Let $\mathcal{L}$ be a list of leaf nodes sorted from left to right
        \For{$L_i \in \mathcal{L}$}
            \State $\mathcal{G},\tau \gets$ \Call{AddLeafInputs}{$\mathcal{G}$,$L_i$,$\tau$} \Comment{Algorithm \ref{alg:add_leaf_inputs}}
            \If{$L_i$ is an action node}
                \State $\mathcal{G} \gets$ \Call{AddLeafOutputs}{$\mathcal{G}$,$L_i$,$\tau$} \Comment{Algorithm \ref{alg:add_leaf_outputs}}
            \EndIf
        \EndFor
        \State $\mathcal{G} \gets$ \Call{AddTemporalEdges}{$\mathcal{G}$,$\tau$} \Comment{Algorithm \ref{alg:add_temporal_edges} (Appendix \ref{app:algorithms})}
        \State \textbf{return} $\mathcal{G}$
    \EndFunction    
    \end{algorithmic}
\end{algorithm}

As the values of state variables can change over time due to the execution of the BT nodes (modelled by $p_i$), it is important to denote the value of each state variable at different times using different variables in the explanation model, similarly to other graphical models that consider a temporal aspect, such as dynamic Bayesian networks \citep{kanazawa1995stochastic}. To facilitate this, we define a correspondence operator $\corresponds$ such that $X^{(t)} \corresponds X$ denotes that the variable $X^{(t)} \in \mathcal{V}$ represents a temporal version of $X \in \mathcal{X}$. To keep track of the temporal versions of each variable, we maintain a mapping $\tau:\mathcal{X}\to\mathbb{Z}^{0+}$ that maps each variable in $\mathcal{X}$ to its latest temporal index.\looseness=-1

To begin, we add an initial version $X^{(0)}$ of each $X\in\mathcal{X}$, representing the state at the beginning of a ``tick'' (Algorithm \ref{alg:cm_from_dk}, line \ref{algl:add_inital_variables_line}). Then, for each leaf node $L_i$, in order from left to right in the BT, we perform two steps - adding incoming edges based on $\mathcal{X}_i$ and outgoing edges based on $\mathcal{Y}_i$.\looseness=-1

In the first step, we add the latest version of $X$ and all of the causal ancestors of $X$, as defined by $\tau$, to the variable set $\mathcal{V}$ for $X\in\mathcal{X}_i$. We relate each of these variables to $r_i$ and, for action nodes, $d_i$ (Algorithm \ref{alg:add_leaf_inputs}). This represents the effect of the state on the return status (and decision) of $L_i$. After adding edges for each leaf node, we increment the values in $\tau$ for the variables in $\mathcal{X}_i$ and their causal ancestors.

\begin{algorithm}
    \caption{Inserting nodes and edges based on $\mathcal{X}_i$}\label{alg:add_leaf_inputs}
    \begin{algorithmic}[1]
        \Function{AddLeafInputs}{$\mathcal{G}$,$L_i$,$\tau$} \funclabel{algf:add_leaf_inputs} \label{algl:add_leaf_inputs}
        \For{$X \in \mathcal{X}_i$}
             \State Add $X^{(\tau(X))}$ to $\mathcal{V}$ if $X^{(\tau(X))} \notin \mathcal{V}$
            \State Let $\mathcal{C}_X$ be the set of ancestors of $X$ in $\mathcal{G}_\mathcal{S}$
            \State $\forall X_C \in \mathcal{C}_X$, add $X_C^{(\tau(X_C))}$ to $\mathcal{V}$ if $X_C^{(\tau(X_C))} \notin \mathcal{V}$
            \For{$X_a \to X_b \in \mathcal{E}_{\mathcal{S}}$}
                \If{$X_a$ or $X_b \in \mathcal{C}(X)\cup\{X\}$}
                    \State Add edge $X_a^{(\tau(X_a))} \to X_b^{(\tau(X_b))}$ to $\mathcal{E}$
                \EndIf
            \EndFor
            \State Add edge $X^{(\tau(X))} \to r_i$ to $\mathcal{E}$
            \If{$L_i$ is an action node}
                \State Add edge $X^{(\tau(X))} \to d_i$ to $\mathcal{E}$
            \EndIf
        \EndFor
        \State Let $\mathcal{C}$ be the set of ancestors of all $X\in\mathcal{X}_i$ in $\mathcal{G}_\mathcal{S}$
        \For{$X \in \mathcal{X}_i\cup\mathcal{C}$}
            \State $\tau(X) \gets \tau(X) + 1$
        \EndFor
        \State \textbf{return} $\mathcal{G}$,$\tau$ \label{algl:add_leaf_inputs_end}
    \EndFunction
    \end{algorithmic}
\end{algorithm}

In the second step, which is exclusive to action nodes, we add the latest version (defined by $\tau$) of every output variable $X\in\mathcal{Y}_i$ to $\mathcal{V}$, and we relate the execution $E_i$ of $L_i$ (and the decision $d_i$ for action nodes) to their values (Algorithm \ref{alg:add_leaf_outputs}). This represents the effect executing $L_i$ (and the particular decision taken by the node) has on the state.

\begin{algorithm}
    \caption{Inserting nodes and edges based on $\mathcal{Y}_i$}\label{alg:add_leaf_outputs}
    \begin{algorithmic}[1]
        \Function{AddLeafOutputs}{$\mathcal{G}$,$L_i$,$\tau$} \funclabel{algf:add_leaf_outputs} \label{algl:add_leaf_outputs}
        \For{$X \in \mathcal{Y}_i$}
            \State Add $X^{(\tau(X))}$ to $\mathcal{V}$ if $X^{(\tau(X))} \notin \mathcal{V}$
            \State Add edge $E_i \to X^{(\tau(X))}$ to $\mathcal{E}$
            \State Add edge $d_i \to X^{(\tau(X))}$ to $\mathcal{E}$
        \EndFor
        \State \textbf{return} $\mathcal{G}$ \label{algl:add_leaf_outputs_end}
    \EndFunction
    \end{algorithmic}
\end{algorithm}

As the counts in $\tau$ are incremented after adding incoming edges to a leaf node, we can ensure that different temporal stages of the state are represented with different variables, and thus the explanation graph remains acyclic. For example, if $L_i$ both reads from and writes to $X$ (i.e. $X\in\mathcal{X}_i\cap\mathcal{Y}_i$), then this update is reflected in the edges $X^{(t)}\to d_i$ and $d_i \to X^{(t+1)}$, rather than a cycle.

Finally, after all leaf nodes have had their input and output edges added, we relate each top-level variable (i.e. variables without parents in $\mathcal{G}_\mathcal{S}$) to themselves temporally, so that a variable can retain its previous value if a leaf node does not alter it (Algorithm \ref{alg:add_temporal_edges} in Appendix \ref{app:algorithms}). We perform this ``temporal linking'' only for nodes without parents, as we assume other state variables have their values determined entirely by their parents. If such a variable also relies on its own previous values, these links can also be added.

\subsubsection{Explanation Model Functions}\label{sec:architecture_causal_model_functions}
Having defined the explanation graph $\mathcal{G}$ in the previous sections, all that remains to define the explanation model $M$ is to define the ranges $\mathcal{R}$ of values and the functions $\mathcal{F}_n(\pmb{v})$, where $\pmb{v}$ is an assignment of variables in $\mathcal{V}$, for each node $n$ in the causal model. The definitions of these functions are described in this section, and depicted in pseudocode form in Algorithm \ref{alg:functions} in Appendix \ref{app:algorithms}. 

Recall that each node in the explanation graph represents either a state variable $X^{(t)}$ or the return status ($r_i$), execution ($E_i$) or decision ($d_i$) of a BT node $\mathcal{T}_i$. For execution variables $E_i$, $\mathcal{R}(E_i) = \{true,false\}$ by definition. The value of $\mathcal{F}_{E_i}$ depends on the relationship of $\mathcal{T}_i$ to its parent and its siblings (Algorithm \ref{alg:execution}, Appendix \ref{app:algorithms}). If $\mathcal{T}_i$ is a root node, then $E_i$ is true, as the root node of the BT is always executed. Otherwise, $\mathcal{T}_i$ must be the child of a composite node $\mathcal{T}_p = Pa(\mathcal{T}_i)$. If $\mathcal{T}_i$ is the left-most child of $Pa(\mathcal{T}_i)$, then it is always executed first and thus $E_i = E_{p}$. Otherwise, the value of $E_i$ depends on the return status of the sibling of $\mathcal{T}_i$ immediately to its left, $\mathcal{T}_j$. If $\mathcal{T}_p$ is a sequence node, then $\mathcal{T}_i$ is only executed if $\mathcal{T}_j$ succeeds, and thus $E_i$ is true only if $r_j = \Success$. If instead $\mathcal{T}_p$ is a fallback node, then $\mathcal{T}_i$ is only executed if $\mathcal{T}_j$ fails, and thus $E_i$ is true only if $r_j = \Failure$.\looseness=-1

For return status variables $r_i$,  $\mathcal{R}(r_i) = \{\Running,\Success,\Failure,\Invalid\}$ by definition (although $\Running$ can be excluded from $\mathcal{R}(r_i)$ in the case of condition nodes). The function $\mathcal{F}_{r_i}$ differs depending on the type of node of $\mathcal{T}_i$ (Algorithm \ref{alg:return}, Appendix \ref{app:algorithms}). In all cases, $r_i = \Invalid$ if $E_i$ is false, by definition. Otherwise, if $\mathcal{T}_i$ is a leaf node, then $r_i$ depends on the state $\pmb{s}$ at the time of node execution, and thus the value of the node is $r_i(\pmb{s})$ ($r_i(d_i,\pmb{s})$ for action nodes). If $\mathcal{T}_i$ is a composite node, its return status depends on the return statuses of its children. For a sequence node, $r_i = \Success$ if all of its children return $\Success$, otherwise taking the value of the first dissenting child. Similarly for a fallback node, $r_i = \Failure$ if all of its children return $\Failure$, otherwise taking the value of the first dissenting child.\looseness=-1

For decision variables $d_i$, the range can be determined by the node's action space, $\mathcal{R}(d_i) = \mathcal{A}_i$. The value of $\mathcal{F}_{d_i}$ can easily be determined as either $\varnothing$ if $E_i$ is false, by definition, or else by evaluating $d_i(\pmb{s})$ (Algorithm \ref{alg:decision}, Appendix \ref{app:algorithms}).\looseness=-1

Finally, for state variables $X^{(t)} \corresponds X \in \mathcal{X}$, the range is already specified by the state model $M_\mathcal{S}$, such that $\mathcal{R}(X^{(t)}) = \mathcal{R}_{\mathcal{S}}(X)$. To determine $\mathcal{F}_{X^{(t)}}$, we note the distinction between internal state variables, whose values can be set by the BT, and external state variables, whose values cannot be set by the BT. The value of an external state variable, whose parents in $\mathcal{G}$ are all state variables, is determined by the corresponding function $\mathcal{F}_{\mathcal{S}X}$ from the state causal model (Algorithm \ref{alg:state_node}, lines \ref{algl:external_state}-\ref{algl:external_state_end}, Appendix \ref{app:algorithms}). For internal state variables, the value is instead determined by $p_i$ for the leaf $L_i$ where $E_i$ and $d_i$ are parents of $X^{(t)}$ in $\mathcal{G}$ (Algorithm \ref{alg:state_node}, lines \ref{algl:internal_state}-\ref{algl:internal_state_end}, Appendix \ref{app:algorithms}), unless $L_i$ was not executed, in which case the value remains the same (i.e. $X^{(t)} = X^{(t-1)}$). 

\subsection{Explanation Generation}\label{sec:architecture_explanation_generation}
Given the episodic memory of a particular execution (Section \ref{sec:architecture_execution}) and the explanation model derived from the structure of a BT and domain knowledge of the state and decision-making (Section \ref{sec:architecture_causal_model}), we can now query the value of any node in the explanation model in order to generate explanations, automatically and in real time, about the BT execution.\looseness=-1

Recall that a query takes the form $\Why(A,B)$, where $A$ and $B$ are mutually exclusive statements (see Section \ref{sec:definitions_counterfactuals}). When querying the explanation model, we set $A = \{V=v\text{ at time }k|V\in\mathcal{V}_Q\}$, where $\mathcal{V}_Q\subseteq\mathcal{V}$ denotes the set of variables we are querying and $V=v \text{ at time }k$ must be true (i.e. $v$ is the value of $V$ at time $k$ in the episodic memory). Similarly, we set $B = \{V\in\mathcal{R}_Q(V)|V\in\mathcal{V}_Q\}$, where $\mathcal{R}_Q(V)\subset\mathcal{R}(V)$ (with $v\notin\mathcal{R}_Q(V)$) denotes a set of untrue (i.e. foil) values of $V$.

For example, consider the example query presented in Section \ref{sec:definitions_counterfactuals}. Supposing that there exists a node in the BT $RepeatOrEnd$ responsible for deciding whether to repeat a sequence or end the task, we can formally rephrase the query as $\Why(d_{RepeatOrEnd}=End\text{ at time }k,d_{RepeatOrEnd}=Repeat\text{ at time }k)$, where $k$ is the time step at which the $RepeatOrEnd$ node executed.

In order to generate an explanation for such a query, we must first recover the state of the BT and state variables at the queried time ($k$). This requires first initialising the explanation model with the initial values stored in the episodic memory and then iterating through the episodic memory from the initial time ($0$) until time $k$, updating values as updates occur. At the end of each ``tick'' prior to time $k$, the state of the tree can be reset and the values of the latest temporal versions of each state variable can be transferred to their initial corresponding versions. The result is that the true values of all  variables in $\mathcal{V}_Q$ (and their ancestors in $\mathcal{G}$) are known at the queried time $k$.

At this point, a set of explanations can be generated by performing a counterfactual search (Algorithm \ref{alg:counterfactual_search}, visualised in Figure \ref{fig:flowchart_counterfactual_search}, Appendix \ref{app:algorithms}). This search is driven by testing interventions (performed via the $Do$ operator) on the ancestors of the variables in $\mathcal{V}_Q$ to see if they elicit any changes in the queried variables that would satisfy the query, following similar approaches by \cite{albini2020relation} and \cite{diehl2022did}. Intuitively, if setting $X_n^{(t)} = x_n^{'}$ instead of $x_n$ results in a change in the variables in $\mathcal{V}_Q$ such that the query is satisfied, then $X_n^{(t)} = x_n$ is a reason for the true values of the variables in $\mathcal{V}_Q$.\looseness=-1

\begin{algorithm}
    \caption{Counterfactual search given query $Q$}\label{alg:counterfactual_search}
    \begin{algorithmic}[1]
        \Function{CounterfactualSearch}{$Q$,$D_{max}$}\funclabel{algf:counterfactual_search} \label{algl:counterfactual_search}
            \State $\mathcal{V}^* \gets \mathcal{V}_Q\cup\{V|V \text{ is an ancestor of }W\in\mathcal{V}_Q\text{ in }\mathcal{G}\}$
            \State $\mathcal{E}^* \gets \{v\to w|v,w\in\mathcal{V}^*, v\to w \in \mathcal{E}\}$
            \State $\mathcal{G}^* \gets (\mathcal{V}^*,\mathcal{E}^*)$
            \State $\mathcal{M}^* \gets (\mathcal{G}^*,\mathcal{R},\mathcal{F})$
            \State $\mathbf{E} \gets \emptyset$, $i \gets 0$
            \While{$\mathbf{E}=\emptyset$ and $i\leq D_{max}$}\label{algl:counterfactual_search_minimality}
                \State $\mathbf{E}^* \gets $\Call{ExplainToDepth}{$Q,\mathcal{M}^*,i$}
                \State Add $\mathbf{E}^*$ to $\mathbf{E}$
                \State $i \gets i+1$
            \EndWhile
            \State \textbf{return} $\mathbf{E}$\label{algl:counterfactual_search_end}
        \EndFunction

        \Function{ExplainToDepth}{$Q,\mathcal{M}^*,i$}\funclabel{algf:explain_to_depth} \label{algl:explain_to_depth}
            \State $\mathbf{E}\gets\emptyset$
            \State $\mathcal{V}_{Search}\gets \mathcal{V}^*\setminus \mathcal{V}_Q$
            \State $C\gets$\Call{Combinations}{$\mathcal{V}_{Search}, i$}
            \For{assignment $c\in C$}
                \State $\mathcal{G}' \gets Do(\mathcal{M}^*,V=v^*~\forall (V,v^*)\in c)$
                \State Let $v'$ denote the value of $V$ in $\mathcal{G}'$
                \If{$v'\in\mathcal{R}_Q(V)~\forall V\in\mathcal{V}_Q$}\label{algl:counterfactual_search_validity}
                    \State $\mathbf{R} \gets \{V=v|(V,v^*)\in c\}$
                    \State $\mathbf{J} \gets \{V=v^*|(V,v^*)\in c\}$
                    \State $\mathbf{K} \gets \{V=v'|V\in\mathcal{V}_Q\}$
                    \State Add $\langle \mathbf{R}, (\mathbf{J},\mathbf{K}) \rangle$ to $\mathbf{E}$
                \EndIf
            \EndFor
            \State \textbf{return} $\mathbf{E}$\label{algl:explain_to_depth_end}
        \EndFunction

        \Function{Combinations}{$\mathcal{V}_{Search}, i$}\label{algl:combinations}
            \State $\text{combinations} \gets \emptyset$
            \For{each subset $\mathcal{V}_i$ of $\mathcal{V}_{Search}$ of size $i$}
                \For{each assignment $V=v^*~\forall V\in\mathcal{V}_i$}
                    \If{$v^*\neq v~\forall V\in\mathcal{V}_i$ in the assignment}
                        \State Add assignment to combinations
                    \EndIf
                \EndFor
            \EndFor
            \State \Return $\text{combinations}$\label{algl:combinations_end}
        \EndFunction
    \end{algorithmic}
\end{algorithm}

More formally, the search operates by first reducing the search space to only consider the ancestors of the variables in $\mathcal{V}_Q$. For every variable $V$ in the reduced variable set $\mathcal{V}^*$, we add every possible setting $V=v^*$ (except $V=v$, the true value of the variable) to a list of potential explanations (Algorithm \ref{alg:counterfactual_search}, lines \ref{algl:combinations} - \ref{algl:combinations_end}). Discretisation of continuous variables is employed to make the search tractable. One by one, we can perform the intervention $Do(V=v^*)$ on the causal model and check if the new values of $V_Q \in \mathcal{V}_Q$ satisfy the query (i.e. if $V=v'$ after intervention, then if $v'\in\mathcal{R}_Q(V)~\forall V\in\mathcal{V}_Q$, the query is satisfied).

If, after intervening $Do(V=v^*)$, the query is satisfied, then we can construct an explanation around $V$ with the reason $\mathbf{R} = \{V=v\}$ and counterfactual $\mathbf{C} = \{(\{V=v^*\},\{V_Q=v'|V_Q\in\mathcal{V}_Q\})\}$. We can read this explanation as ``$V_Q = v_Q~\forall V_Q\in\mathcal{V}_Q$ because $V=v$. If instead $V=v^*$, then $V_Q = v'~\forall V_Q \in \mathcal{V}_Q$.''.

If no explanations using only a single explanation variable are found, we can perform a ``deeper'' search by considering interventions on two or more variables simultaneously, until either an explanation is found or the search has reached a maximum depth. At the end of the process, the algorithm will return all explanations consisting of $i$ variables, where $i$ is the minimum depth for which explanations are found.

With such a search, we can return a set of explanations for any query regarding the execution $E_i$, return status $r_i$ or decision $d_i$ of a BT node, or the value of a state variable $X^{(t)}$, the accuracy of which is limited only by the completeness of the explanation model, which is determined by the domain knowledge used to construct it, and the granularity of the discretisation in the case of continuous variables.\looseness=-1

Once explanations have been provided, a user could make further ``follow-up'' queries based on the explanations received. These queries can be used to clarify an explanation, or dig deeper towards a root cause (e.g. if the user asks $\Why(A,B)$ and receives an explanation $\langle \{X=x\}, \{(X=x',B)\} \rangle$, a natural follow-up question could be $\Why(X=x,X=x')$. Similarly, follow-up questions allow a user to extend the time window of explanations beyond a single BT ``tick'', by recognising that the value of $X_i^{(0)}$ at tick $t$ is the same as $X_i^{(\tau(X_i))}$ at tick $t-1$. Thus for explanation $\langle \{X_i^{(0)}=x\}, \{(X_i^{(0)}=x',B)\} \rangle$, the follow-up can be $\Why(X_i^{(\tau(X_i))}=x,X_i^{(\tau(X_i))}=x')$ in the context of the previous tick.\looseness=-1

\subsection{Properties}\label{sec:architecture_properties}
Having presented the entire approach to building an explanation model of the BT and state and using it to generate counterfactual explanations, we now discuss some properties of this approach.

\cite{guidotti2022counterfactual} identifies a number of desirable properties for counterfactual explanations, several of which achieved through the design of our approach. For example, a counterfactual explanation is considered \textit{valid} if the counterfactual situation proposed does indeed result in the foil decision. All explanations returned by Algorithm \ref{alg:counterfactual_search} are \textit{valid} by construction, as only explanations that satisfy the query are included in the explanation set (Algorithm \ref{alg:counterfactual_search}, line \ref{algl:counterfactual_search_validity}). Similarly, our explanations are \textit{minimal} - meaning the number of variable value changes between the real assignment and the counterfactual is minimal - as the search incrementally increases the number of variables changed (the ``depth'') until at least one explanation is found (Algorithm \ref{alg:counterfactual_search}, line \ref{algl:counterfactual_search_minimality}).\looseness=-1

Likewise, the explanation set is maximally \textit{diverse} - meaning the returned counterfactuals are as different from each other as possible - as the search covers all combinations of a given depth and thus changes to all potential variable combinations are considered (Algorithm \ref{alg:counterfactual_search}, lines \ref{algl:combinations}-\ref{algl:combinations_end}). The explanations also respect the \textit{causal} relationships in the BT and the environment, as this information is provided by the structure of the BT and the domain knowledge passed to the system. Ultimately, the causal fidelity is limited by the fidelity of the domain knowledge, as discussed in Section \ref{sec:case_study_domain_knowledge}.\looseness=-1

Other properties such as \textit{plausibility}, \textit{actionability} and \textit{discriminative power} identified by \cite{guidotti2022counterfactual} are not accounted for by our approach, although they could easily be incorporated by further restricting the search space of explanations. For example, \textit{discriminative power} could be improved by only considering variables for which all changes result in a valid counterfactual outcome, as discussed by \cite{albini2020relation}. \textit{Plausibility} could be improved by encoding some notion of which state configurations are more probable than others. It is worth noting that the property of causality already overlaps somewhat with plausibility, and that any interventions on the explanation model in our approach will have plausible downstream effects, even if the intervention itself may be implausible.\looseness=-1

\subsubsection{Complexity of the Explanation Model}\label{sec:complexity_explanation_model}
Based on the node insertions in Algorithms \ref{alg:cm_from_bt} and \ref{alg:cm_from_dk}, we note that the number of nodes in the explanation model grows with the number of nodes in the BT (especially leaf nodes), the number of variables in the state model ($|\mathcal{X}_{\mathcal{S}}|$) and the sizes of the input and output sets ($|\mathcal{X}_i|$ and $|\mathcal{Y}_i|$), as well as the number of causal ancestors of the input variables in the state model. 

More formally, let $\mathcal{T}$ denote the set of BT nodes, with $\mathcal{L} \subset \mathcal{T}$ denoting the set of leaf nodes and $A \subseteq \mathcal{L}$ denoting the set of action nodes. Assume a worst-case scenario in which every state variable is both read and potentially written to by every BT node. In other words, we assume that $\mathcal{X}_i = \mathcal{X}_{\mathcal{S}}\text{ }\forall L_i \in \mathcal{L}$ and that $\mathcal{Y}_i = \mathcal{X}_{\mathcal{S}}\text{ }\forall \mathcal{T}_i \in \mathcal{T}$. Given that Algorithm \ref{alg:cm_from_bt} always inserts $2|\mathcal{T}| + |A|$ nodes in the explanation model, and Algorithm \ref{alg:cm_from_dk} inserts a node into the model every time a state variable appears in $\mathcal{X}_i$ (or an ancestor of such a variable in $\mathcal{G}_\mathcal{S}$) or in $\mathcal{Y}_i$, then the number of nodes in the explanation model grows asymptotically with $\mathcal{O}(2|\mathcal{T}| + |A| + |\mathcal{X}_{\mathcal{S}}|(|\mathcal{L}|+|A|))$ which simplifies to $\mathcal{O}(|\mathcal{T}|+|\mathcal{X}_{\mathcal{S}}|\times|\mathcal{L}|)$, and is in turn bounded above by the simpler $\mathcal{O}(|\mathcal{X}_{\mathcal{S}}|\times|\mathcal{T}|)$.

To determine how the number of edges in the explanation model scales with BT and state model size, we once again assume the worst-case scenario of $\mathcal{X}_i = \mathcal{Y}_i = \mathcal{X}_{\mathcal{S}}$, and also assume that the state model $\mathcal{M}_\mathcal{S}$ contains the maximum number of edges, which for a directed acyclic graph is $\frac{1}{2}|\mathcal{X}_{\mathcal{S}}|(|\mathcal{X}_{\mathcal{S}}|-1)$. From Algorithm \ref{alg:cm_from_bt}, we observe that two edges are added for each node (except the root of the tree, for which only one is added), and a further two edges are added for each action node. Additionally, one edge is added for each parent-child relationship in the tree, which is always exactly $|\mathcal{T}|-1$ edges due to the properties of trees. From Algorithm \ref{alg:cm_from_dk}, we observe that for each leaf node, edges are added for every variable in $\mathcal{X}_i$, replicating the edges in $\mathcal{E}_{\mathcal{S}}$ for each temporal version of the state variables. Edges are also added for each action node proportional to the number of input and output variables. Edges linking temporal versions of top-level state variables (of which there is only one in a maximally connected DAG) are added for each action node potentially changing that variable's value. Taken altogether, the number of edges in the explanation model grows asymptotically with $\mathcal{O}(3|\mathcal{T}| + 2|A| + 0.5|\mathcal{X}_{\mathcal{S}}|^{2}\times|\mathcal{L}| + 0.5|\mathcal{X}_{\mathcal{S}}|\times|\mathcal{L}| + |A|\times|\mathcal{X}_{\mathcal{S}}|)$, which simplifies to $\mathcal{O}(|\mathcal{X}_{\mathcal{S}}|^{2}\times|\mathcal{L}|+|\mathcal{T}|)$, and is in turn bounded above by the simpler $\mathcal{O}(|\mathcal{X}_{\mathcal{S}}|^{2}\times|\mathcal{T}|)$.

These analyses highlight the important role the size of the BT and the state model have in determining the size of the explanation model, which ultimately impacts the runtime complexity of the counterfactual search (as we expand on in Section \ref{sec:complexity_counterfactual_search}).

\subsubsection{Complexity of the Counterfactual Search}\label{sec:complexity_counterfactual_search}

We now examine the runtime complexity of the counterfactual search algorithm (Algorithm \ref{alg:counterfactual_search}) for an explanation model with $N$ nodes. We begin by observing that the expensive operations that occur during the execution of the search are the application of the $Do$ operator on the reduced sub-model $\mathcal{M}^*$, as well as checking to see if the intervention has satisfied the query $Q$. Both operations occur once for each combination of variable assignments (the set of which is labelled $C$) considered and for each level of depth considered up to the maximum depth $D_{max}$.

Applying the operation $Do(X=x)$ deletes the incoming edges to $X$, changes the value to $x$ and then propagates the effects of the change to every descendant of $X$ in $\mathcal{M}^{*}$. First of all, let us assume the worst-case scenario in which the queried variable has as many ancestors as possible. To create this upper bound, we assume $\mathcal{M^*}=\mathcal{M}$, and thus every node in the explanation model must be considered in the counterfactual search. We also assume a worst-case cost of $\mathcal{O}(F)$ for each function in $\mathcal{F}$. In the absolute worst case, a given $Do(X=x)$ necessitates $N-1$ function executions, and thus runtime complexity for the $Do$ operator in this scenario is $\mathcal{O}(NF)$.

The search considers multiple interventions on a single variable, one for each possible value the variable can take (discretised over a range, in continuous cases). In a worst-case simplification, we assume each node in the explanation model has $R_{max}$ possible interventions that must be considered. Therefore, at a given depth $D$, there are $\binom{N}{D}R_{max}^D$ combinations of assignments, and thus, taken together with the runtime complexity of the $Do$ operator, we have a runtime complexity of $\mathcal{O}(NF\binom{N}{D_{max}}R_{max}^{D_{max}})$. Given that $N$ grows asymptotically bounded above by $\mathcal{O}(|\mathcal{X}_{\mathcal{S}}|\times|\mathcal{T}|)$, we can express the counterfactual search runtime complexity in terms of BT and state model size as $\mathcal{O}((|\mathcal{X}_{\mathcal{S}}|\times|\mathcal{T}|)F\binom{|\mathcal{X}_{\mathcal{S}}|\times|\mathcal{T}|}{D_{max}}R_{max}^{D_{max}})$

It is worth recalling that the runtime complexity calculated above assumes the worst-case scenario, including for the state and explanation model structures, possible variable values and BT node input/output sets. In particular while a high value of $D_{max}$ may be set, early stopping at lower depths greatly reduces runtime. Additionally, the complexity $\mathcal{O}(F)$ is often $\mathcal{O}(1)$ in typical BT implementations (for example, in nodes where preconditions are checked or flags are set). The empirical analysis in Section \ref{sec:evaluation} shows runtime to be low, even for the larger and more complex BT examined in Section \ref{sec:evaluation_serial_recall}.\looseness=-1

%%%
%%% CASE STUDY
%%%

\section{Case Study}\label{sec:case_study}
\begin{figure}[]
    \centering
    \includegraphics[width=0.4\linewidth]{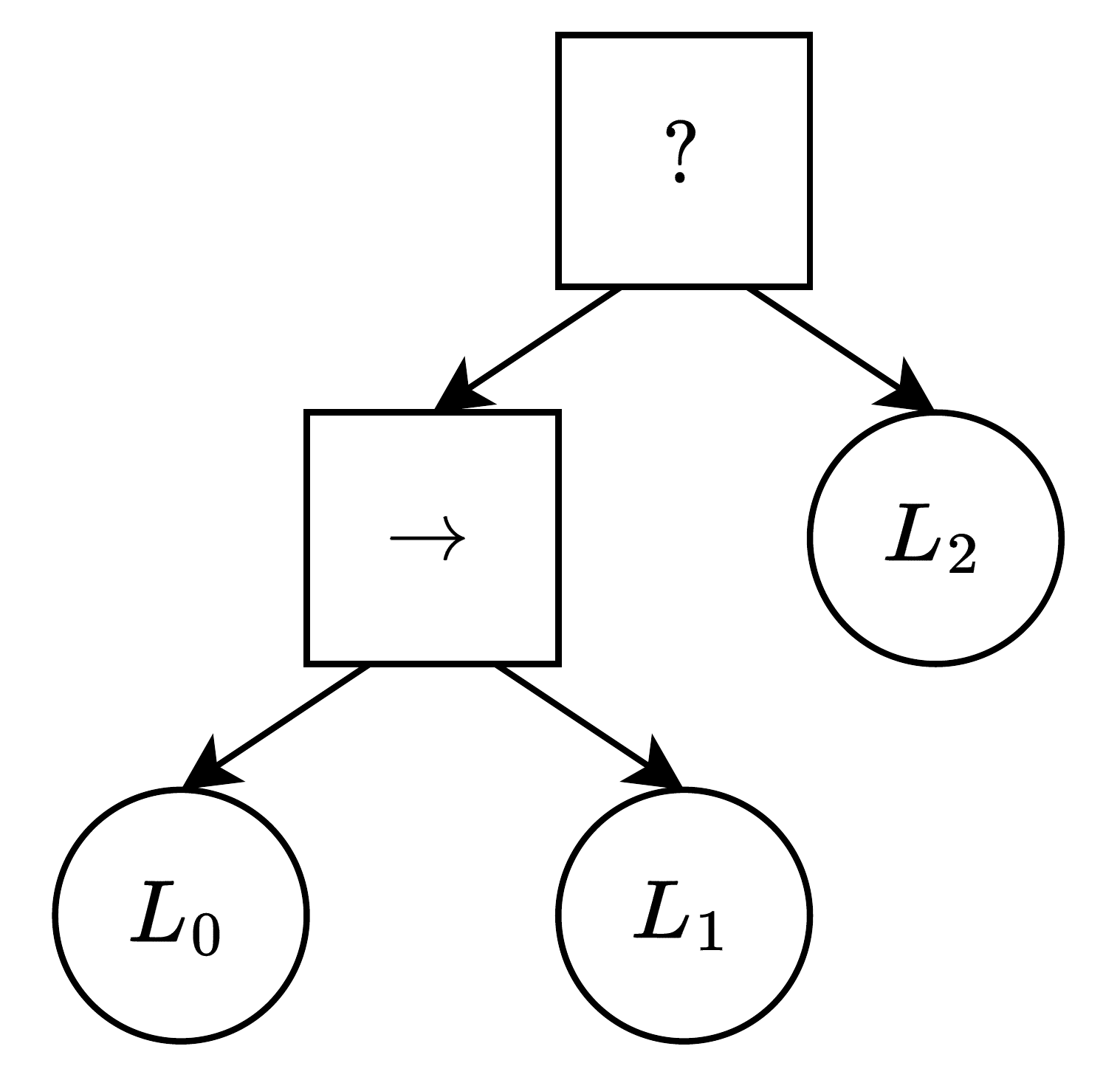}
    \caption{\footnotesize
    An example BT with a fallback node $\mathcal{T}_?$, sequence node $\mathcal{T}_{\rightarrow}$, condition node $L_0$ and action nodes $L_1$ and $L_2$. The leaf nodes in this tree have the following input, output and action sets: $\mathcal{X}_0 = \{X_a\}$, $\mathcal{X}_1 = \{X_a, X_c\}$, $\mathcal{X}_2 = \{X_d\}$, $\mathcal{Y}_0 = \emptyset$, $\mathcal{Y}_1 = \{X_b\} = \mathcal{Y}_2$, $\mathcal{A}_0 = \emptyset$, $\mathcal{A}_1 = \{a_0,a_1\}$ , $\mathcal{A}_2 = \{a_2,a_3\}$}
    \label{fig:case_study_bt}
\end{figure}

To demonstrate how our architecture is able to generate explanations for a wide range of queries, we provide some illustrative examples using a simple toy behaviour tree (depicted in Figure \ref{fig:case_study_bt}) and state model (in this case, identical to the example in Figure \ref{fig:state_graph_example}, with the assumption that each variable is a boolean statement).\looseness=-1

\begin{figure}[]
    \centering
    \includegraphics[width=0.5\linewidth]{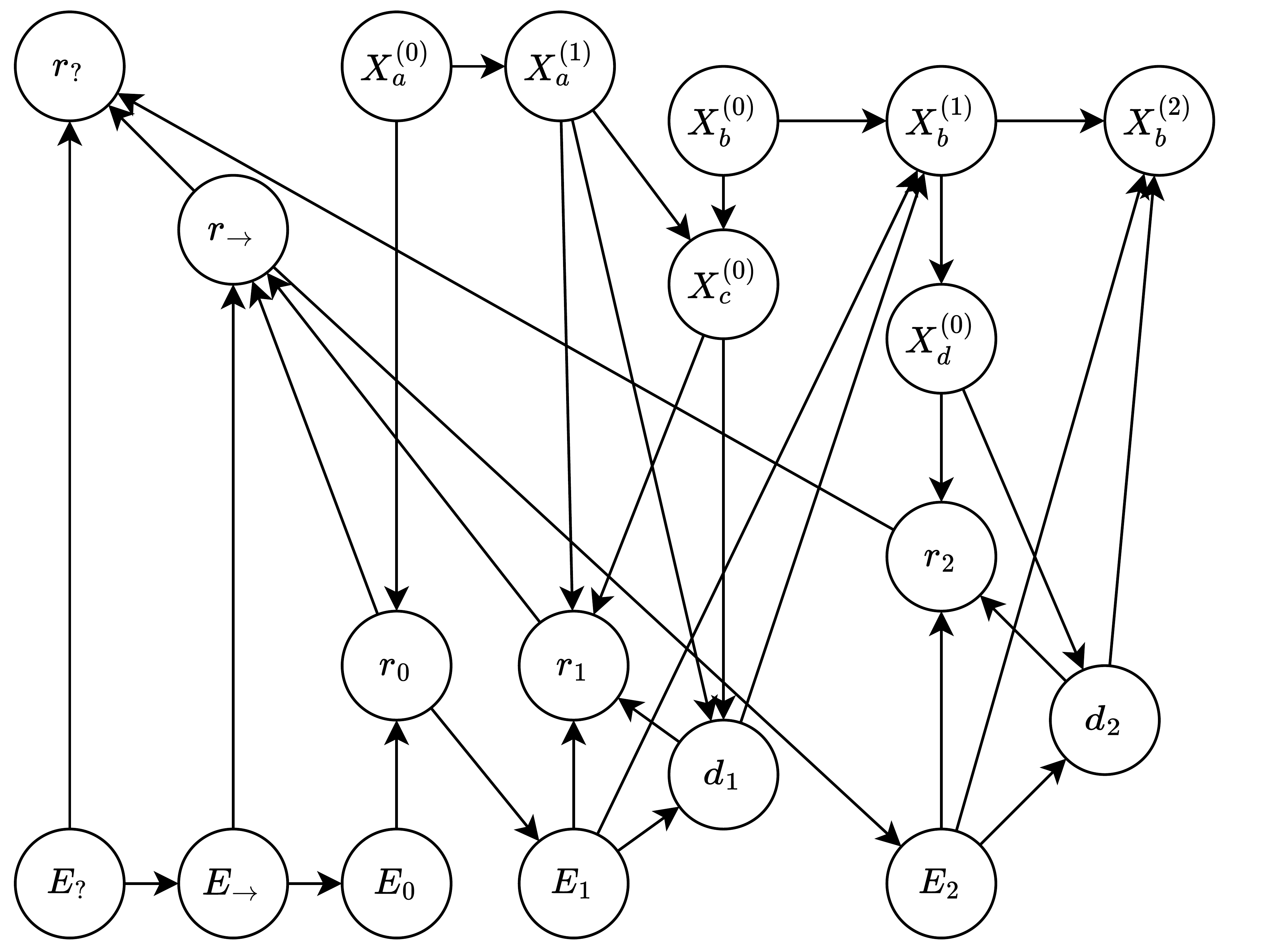}
    \caption{\footnotesize
    The complete explanation model for the behaviour tree in Figure \ref{fig:case_study_bt} and the state graph in Figure \ref{fig:state_graph_example}, constructed using Algorithm \ref{alg:causal_graph}.}
    \label{fig:case_study_cm}
\end{figure}

Regardless of the query, we first need to construct an explanation model using the algorithms presented in Section \ref{sec:architecture_causal_model}. In this case, the resulting causal model is depicted in Figure \ref{fig:case_study_cm}. Using this model, we can query the return statuses, execution and decisions of BT nodes as well as the values of state variables.

\subsection{Step-by-Step Example}\label{sec:case_study_example}

For example, suppose that the condition node $L_0$ has been implemented in such a way as to return $\Success$ if $X_a$ is true and $\Failure$ otherwise. If we were to set $X_a = False$ and execute the BT, finding that $r_0 = \Failure$ at time $1$, a valid query may be to ask $\Why(r_0 = \Failure \text{ at time }1, r_0 = \Success)$. We now provide a step-by-step description of Algorithm \ref{alg:counterfactual_search} for this particular query to illustrate how it arrives at an explanation.

Firstly, the algorithm identifies a subgraph $\mathcal{G}^*$ containing only the query variables $\mathcal{V}_Q$ and their ancestors in $\mathcal{G}$. Examining the explanation model in Figure \ref{fig:case_study_cm}, we arrive at the set $\mathcal{V}^* = \{E_?, E_{\rightarrow}, E_0, X_a^{(0)}, r_0\}$ for query variable $\mathcal{V}_Q = \{r_0\}$. The next step is to identify all possible interventions in the search space $\mathcal{V}^*\setminus \mathcal{V}_Q$, which, given that each variable in the set happens to be a boolean statement, is $C = \{E_? = False, E_{\rightarrow} = False, E_0 = False, X_a^{(0)} = True\}$. Note how the true values of these variables (e.g. $E_0 = True$ or $X_a^{(0)} = False$) are omitted from the search space.\looseness=-1

For each statement $V = v^*$ in $C$, we can perform the intervention $Do(V=v^*)$ to see if the resulting values in $\mathcal{M}^*$ satisfy our query. In this case, to satisfy the query, the value of $r_0$ after the intervention must be $\Success$. We are able to obtain the counterfactual value of $r_0$ as we are able to run the node with new input variables, thus evaluating the function $r_0(\pmb{s})$. Attempting each intervention, we note that $Do(E_?=False)\implies r_0 = \Invalid$, $Do(E_{\rightarrow}=False)\implies r_0 = \Invalid$, $Do(E_0=False)\implies r_0 = \Invalid$ and $Do(X_a^{(0)}=True)\implies r_0 = \Success$. As only the intervention $Do(X_a^{(0)}=True)$ satisfies the query, we arrive at a single explanation $\langle \mathbf{R} = \{X_a^{0} = False\}, \mathbf{C} = (\{X_a^{0} = True\},\{r_0 = \Success\}) \rangle$, which in natural language could be converted to ``$L_0$ failed because $X_a^{(0)}$ was false. If $X_a^{(0)}$ was true, $L_0$ would have succeeded.''. Of course, this is a simple example as $r_0$ has only a few ancestors in the causal model. The query $\Why(r_{\rightarrow} = \Failure \text{ at time }t, r_? = \Success)$, for example, would result in 8 possible explanations due to the much larger set of ancestors of $r_{\rightarrow}$.

The same algorithm can be applied to query any variable in the explanation model. Example queries and corresponding explanations are given in Table \ref{tab:example_queries}. In particular, note the explanations provided for the first example, $\Why(E_0 = True\text{ at time }1, E_0 = False)$. These explanations note that the execution $L_0$ is the result of the chain of executions of its ancestors, $\mathcal{T}_?$ and $\mathcal{T}_{\rightarrow}$. 

\begin{table*}[]
\caption{Some example queries and their corresponding explanations, given the explanation model in Figure \ref{fig:case_study_cm}. For this case study, we have $L_2$ select $a_2$ if $X_d = True$, otherwise selecting $a_3$. We also have $X_c = X_a \cap X_b$ and $X_d = \neg X_b$. Note that the natural language queries and explanations (only reasons, for the sake of brevity) are included purely for the purposes of legibility. Our architecture takes in a formal query as input and produces the formal explanation set as output.}
\label{tab:example_queries}
\resizebox{\textwidth}{!}{%
\begin{tabular}{llll}
\textbf{Natural Language Query} & \textbf{Formal Query} & \textbf{Explanation Set}  & \textbf{Natural Language Explanation}\\ \hline
\multirow{2}{*}{``Why did $L_0$ execute?''} & \multirow{2}{*}{\twolinetablecell[]{$\Why(E_0 = True\text{ at time }1,$, \\$ E_0 = False)$}} & $e_1 = \langle E_{\rightarrow} = True, (E_{\rightarrow} = False, E_0 = False) \rangle$  & ``...because $\mathcal{T}_{\rightarrow}$ executed...'' \\
 &  &  $e_2 = \langle E_? = True, (E_? = False, E_0 = False) \rangle$ & ``...because $\mathcal{T}_{?}$ executed...'' \\ \hline
\multirow{2}{*}{``Why didn't $L_1$ execute?''} & \multirow{2}{*}{\twolinetablecell[]{$\Why(E_1 = False\text{ at time }2,$\\$ E_1 = True)$}} & $e_1 = \langle r_0 = \Failure, (r_0 = \Success, E_1 = True) \rangle$ & ``...because $\mathcal{T}_{0}$ failed...'' \\
 &  &  $e_2 = \langle X_a^{(0)} = False, (X_a^{(0)} = True, E_1 = True) \rangle$  & ``...because $X_a^{(0)}$ was false...''\\ \hline
\multirow{3}{*}{\twolinetablecell[]{Why did $L_2$ choose $a_2$ \\ instead of $a_3$?}} & \multirow{3}{*}{\twolinetablecell[]{$\Why(d_2 = a_2\text{ at time }3,$\\$d_2 = a_3)$}} & $e_1 = \langle X_b^{(0)} = False, (X_b^{(0)} = True, d_2 = a_3) \rangle$  & ``...because $X_b^{(0)}$ was false...''\\
 &  & $e_2 = \langle X_b^{(1)} = False, (X_b^{(1)} = True, d_2 = a_3) \rangle$  & ``...because $X_b^{(1)}$ was false...''\\
 &  & $e_3 = \langle X_d^{(0)} = True, (X_d^{(0)} = False, d_2 = a_3) \rangle$  & ``...because $X_d^{(0)}$ was true...''\\ \hline
``Why was $X_c$ false?'' & \twolinetablecell[]{$\Why(X_c = False\text{ at time }2,$\\$X_c = True)$} & \twolinetablecell[]{$e_1 = \langle X_a^{(1)} = False \cap X_b^{(0)} = False,$ \\ $(X_a^{(1)} = True \cap X_b^{(0)} = True, X_c^{(0)} = True) \rangle$}   & \twolinetablecell[]{``...because $X_a^{(1)}$ was false \\ and $X_b^{(0)}$ was false...''} 
\end{tabular}}
\end{table*}

\subsection{Comparison with Other Behaviour Tree Methods}\label{sec:comparison_with_literature}
Using the BT depicted in Figure \ref{fig:case_study_bt}, we now compare our approach to that of \cite{han2021building} to illustrate how our approach extends both the types of causal queries that can be provided to the system as well as the types of reasons that can be included in the explanation. In addition to other types of queries which are not within the scope of our work, the method of \cite{han2021building} is capable of answering the question ``Why are you doing this?''. Within our framework of contrastive queries, this question can be expressed as the formal query ``$\Why(E_i=True,E_i=False)$'', where $E_i$ is the currently executing node. For this example, suppose the node of interest is $L_1$ and that we set $X_a = True$, such that $L_0$ returns $\Success$ and thus $E_i = E_1 = True$. To answer the query, the method of \cite{han2021building} would return a non-decorator ancestor of $L_1$, in this case $\mathcal{T}_{\rightarrow}$ (assuming the two nodes have different names). Interpreting this causally, we would have the reason $E_{\rightarrow} = True$. Our method, on the other hand, would also include $E_? = True$, $E_0 = True$, $r_0 = \Success$ and $X_a^{(0)} = True$ as reasons, and would provide counterfactuals for each, demonstrating that our approach allows for more types of reasons to be considered in the counterfactual search.

It is worth noting that the method of \cite{ogren2023creating} would also include these reasons in the explanation, assuming that the BT in Figure \ref{fig:case_study_bt} is transformed to a suitable ``Make-sure'' BT. However, both this method and that of \cite{han2021building} are limited in the types of causal queries which the system can respond to. As demonstrated in Table \ref{tab:example_queries}, our method can answer queries about decisions ($d_i$), return statuses ($r_i$) and state variable values in addition to execution statuses ($E_i$). Our method can answer queries about events which did not occur (e.g. ``$\Why(E_1 = False, E_1 = True)$''), and can consider specific contrastive foils when generating explanations (e.g. ``$\Why(d_2=a_2,d_2=a_3)$'').

\subsection{The Importance of Domain Knowledge}\label{sec:case_study_domain_knowledge}
Thus far we have considered explanations generated from the complete explanation model, built from the structure of the BT (Section \ref{sec:architecture_causal_model_bt_structure}) as well as domain knowledge in the form of a state model and input/output/action sets for each BT node (Section \ref{sec:architecture_causal_model_domain_knowledge}). While the BT structure is always known, the domain knowledge used to construct the explanation model may be incomplete or incorrect (e.g. due to model simplifications or misspecifications). We now examine the effect of missing or additional edges in the explanation model to highlight the importance of this domain knowledge.\looseness=-1

Firstly, consider that, for a given leaf node $L_i$, edges between nodes representing temporal versions of state variables and $r_i$, $d_i$ and $E_i$ are determined by the sets $\mathcal{X}_i$, $\mathcal{Y}_i$ and $\mathcal{A}_i$, representing the inputs, outputs and action space of the node respectively. If these sets were not known, we would not be able to effectively link changes to the state to the decisions and outcomes of BT nodes (and vice-versa). In order to be able to include these relationships, we would have to add the edges $X^{(t)} \rightarrow r_i$, $X^{(t)} \rightarrow d_i$, $d_i \rightarrow X^{(t+1)}$ and $E_i \rightarrow X^{(t+1)}$ $\forall X \in \mathcal{X}$ to the graph, greatly increasing the number of edges and thus the size of the search space when performing the counterfactual search. Thus, the inclusion of $\mathcal{X}_i$, $\mathcal{Y}_i$ and $\mathcal{A}_i$ serve to reduce the size and complexity of the counterfactual search.

If instead these sets are present but are incorrect, such that elements that should be contained in these sets are excluded, then certain explanations may not be findable by the counterfactual search. For example, consider the explanation model in Figure \ref{fig:case_study_cm}. If we were to set $\mathcal{Y}_1 = \emptyset$ rather than $\mathcal{Y}_1 = \{X_b\}$, then the edges $d_1 \rightarrow X_b^{(1)}$ and $E_1 \rightarrow X_b^{(1)}$ would be excluded from the graph. Therefore, any interventions made to $r_1$, $d_1$, and their ancestors (which are not also ancestors to $X_b^{(1)}$) would not be considered in the counterfactual search and thus would not appear in explanations querying $X_b^{(1)}$ and its descendants. This highlights a trade-off between the search efficiency introduced by including the decision-making knowledge and the possibility of missing explanations due to the inclusion of misspecified input, output and action sets.

Secondly, consider that all edges $X_m^{(t_i)} \rightarrow X_n^{(t_j)}$ where $m \neq n$ are derived from knowledge introduced by the state model $\mathcal{M}_\mathcal{S}$. If the state model is missing or incomplete, then certain explanations would be excluded from the counterfactual search results. For example, returning to the model in Figure \ref{fig:case_study_cm}, if the edge $X_b \rightarrow X_d$ were to be excluded, then the edge $X_b^{(1)} \rightarrow X_d^{(0)}$ would not be present in the explanation model. Thus, while $X_d^{(0)}$ would potentially appear in explanations querying $r_2$, $X_b^{(1)}$ would not, and neither would its ancestors, such as $E_1$ and $d_1$.\looseness=-1

In summary, the accuracy and completeness of the explanations generated by a counterfactual search over the model are limited by the accuracy and completeness of the model (as is true for any approach which relies on causal models), which are in turn determined by the fidelity of the domain knowledge used to construct the model.

%%%
%%% EVALUATION
%%%

\section{Evaluation}\label{sec:evaluation}
Having provided simple illustrative examples to demonstrate the effectiveness of the algorithms proposed in Section \ref{sec:architecture}, we now focus on a more systematic evaluation of the explanations generated by our architecture. There are many different approaches to evaluating explanations. \cite{doshi2017towards} identify three broad types of evaluation depending on the evaluation context and the presence of human evaluators. \textit{Application-} and \textit{human-grounded} approaches both involve user studies, the primary difference being the fidelity to the intended application domain. On the other hand, \textit{functionally-grounded} approaches define an appropriate proxy measure which can be evaluated automatically. Evaluations both with and without user studies can be insightful, depending on the particular property of the explanations being measured \citep{nauta2023anecdotal}. Given that explanation generation is only one component of the broader explainability interaction (which also includes converting a human's request into a formal explanation query and communicating a formal explanation in an effective manner~\citep{anjomshoae2019explainable,matarese2021user}), we opt for a functionally-grounded approach in order to validate our architecture in isolation.\looseness=-1

Several desirable properties have been identified both for explanations in general \citep{nauta2023anecdotal} and counterfactual explanations in particular \citep{guidotti2022counterfactual}, as discussed in Section \ref{sec:architecture_properties}. We focus our evaluations on measuring the central property of \textit{correctness} (otherwise known as \textit{fidelity}), which describes the faithfulness of the explanations to the model being explained. \cite{nauta2023anecdotal} identify several metrics for evaluating correctness, including the \textit{controlled synthetic data check} - which operates by creating a synthetic dataset such that the model under evaluation reasons in a certain way and comparing the explanations against this ground-truth reasoning - and the \textit{single/incremental deletion} - which operates by perturbing or removing one or more features and comparing the size of the effect on the model to the importance score given by a feature importance explanation. Such approaches are aimed at evaluating explanations of black box models learned from data, the absence of which limits their applicability to our architecture.\looseness=-1

To measure correctness and demonstrate that our counterfactual search reflects the reasoning of the BT and the dynamics of the environment (as captured by the state model), we therefore introduce the \textit{target recovery} metric, inspired other evaluation approaches that perturb the initial system input, such as the \textit{controlled synthetic data check} identified by \cite{nauta2023anecdotal}. Here we outline the procedure for calculating the metric.

We begin by labelling an arbitrary initial state configuration as the ``default'' initialisation, resulting in a corresponding ``default behaviour'' by the BT. We note the particular sequence of return statuses, executions and actions produced by the default behaviour. We then perturb the initial state by altering the value of a single state variable, resulting in an ``altered behaviour'' by the BT. We denote this perturbed variable as the ``target variable'' If the default and altered behaviours consist of an identical sequence of node return statuses, executions and decisions (as might occur if a small change was made, for example), then we cannot apply the evaluation. If, however, there is a deviation between the altered behaviour and the default behaviour, we can construct a query from that difference. For example, suppose that the default behaviour results in the event $A = a$ at time $t$, and that the altered behaviour is identical to the default behaviour for times $\{0 ,...,t-1\}$ but has $A=a'$ at time $t$. We can therefore pose the query $\Why(A=a'\text{ at time }t,A=a)$ in the context of the altered behaviour and, using Algorithm \ref{alg:counterfactual_search}, we can generate a set of explanations $\langle \mathbf{R},\mathbf{C} \rangle$. We consider the \textit{target recovery} to be successful if the target variable is contained in the reason set $\mathbf{R}$. For example, if the perturbation is $X=x'$, where the default initialisation has $X=x$, the \textit{target recovery} is successful if $\{X=x'\} \in \mathbf{R}$.

If, over a number of different perturbations and executions, we obtain a high \textit{target recovery} rate (\textit{TRR}), then this suggests our algorithm is capable of accurately recovering the cause of a difference in behaviour.

All experiments described in the following sections have been executed on a device running Ubuntu 22.04.5 LTS with python 3.12.6, 62 GiB RAM, 13th Gen Intel Core i9-13900 CPU and NVIDIA GeForce RTX 4090 24 GiB GPU.

\subsection{Random Domains}\label{sec:evaluation_random}

To demonstrate that our explanation generation method is capable of generating explanations that accurately reflect the decision-making of a wide class of BTs, we evaluate the \textit{target recovery} metric across a number of randomly generated BTs and state dynamics.\looseness=-1

The random BTs are determined by the number of leaf nodes (in our experiment, either 2, 4 or 8). Random tree structures are generated by incrementally adding nodes as children to existing nodes until the number of leaves matches the specified parameter and every composite node has two or more children. Composite nodes are then randomly assigned as either a sequence or fallback node, while every leaf node is randomly assigned an input, output and action set. Random execution functions are generated for each node that are coherent with these sets. The random state graphs are determined by the number of state variables (4, 8 or 12) and a connectivity parameter (0, 0.25, 0.5, 0.75, 1) which determines the number of edges. For each graph, half of the nodes are designated ``top-level'' nodes and have no parents. Each variable is a boolean statement, whose value is a random composition (using ``and'' and ``or'' operators) of its parents in the state graph.

For each combination of number of leaves, number of state variables and connectivity parameter, 10 BT-state graph combinations are generated using different seeds, for a total of 450 unique combinations. For each combination, a random assignment of state variable values are designated as the ``default''. Additionally, we perturb each of the designated ``top-level'' state variables (by flipping the boolean value) to produce a number of altered executions. For each combination, we compare the altered executions with the default one to evaluate the \textit{target recovery} metric. In total, 1800 comparisons are made. For each assignment of BT and state parameters, the \textit{TRR} and runtime statistics are given in Table \ref{tab:random_results_target_runtime}. Additionally, we provide the size of the explanation model (i.e. the number of nodes in the causal model), the number of counterfactual candidates considered in the counterfactual search (i.e. the interventions considered when producing the explanation set) and the number of explanations provided when explaining a given altered execution in Table \ref{tab:random_results_numbers}.

\begin{table*}[]
\caption{Results for target recovery experiments with randomly generated state graphs and behaviour trees. For the sake of compactness, the connectivity parameter is omitted. In each run either the target variable is successfully recovered or there is no difference in execution. In addition to target recovery rate (\textit{TRR}), we also show the runtime, which is the time taken to generate the explanation set after receiving a query.}
\label{tab:random_results_target_runtime}
\resizebox{\textwidth}{!}{%
\begin{tabular}{cc|cc|c|ccc}
\multicolumn{2}{c|}{\textbf{BT/State Parameters}} & \multicolumn{2}{c|}{\textbf{Number of Runs}} & \multirow{2}{*}{\textbf{TRR}} & \multicolumn{3}{c}{\textbf{Runtime (s)}}   \\
\textbf{No. State Vars}      & \textbf{No. BT Leaves}     & \textbf{Target Found}    & \textbf{No Difference}    &                                       & \textbf{Min.}   & \textbf{Max.}   & \textbf{Mean (Std)}      \\ \hline
4                   & 2                 & 49              & 51               & 1                                     & 0.0008 & 0.0076 & 0.0026 (0.0019) \\
4                   & 4                 & 48              & 52               & 1                                     & 0.0008 & 0.0166 & 0.0032 (0.0042) \\
4                   & 8                 & 42              & 58               & 1                                     & 0.0012 & 0.0786 & 0.0170 (0.0204) \\
8                   & 2                 & 76              & 124              & 1                                     & 0.0011 & 0.0263 & 0.0053 (0.0054) \\
8                   & 4                 & 79              & 121              & 1                                     & 0.0011 & 0.0803 & 0.0089 (0.0141) \\
8                   & 8                 & 93              & 107              & 1                                     & 0.0017 & 0.2716 & 0.0355 (0.0573) \\
12                  & 2                 & 81              & 219              & 1                                     & 0.0012 & 0.0655 & 0.0132 (0.0132) \\
12                  & 4                 & 104             & 196              & 1                                     & 0.0012 & 0.241  & 0.0278 (0.0473) \\
12                  & 8                 & 115             & 185              & 1                                     & 0.0026 & 0.3709 & 0.0775 (0.0845) \\ \hline
All                 & All               & 687             & 1113             & 1                                     & 0.0008 & 0.3709 & 0.0266 (0.0517)
\end{tabular}}
\end{table*}

\begin{table*}[]
\caption{Additional results for the experiments conducted with randomly generated state graphs and behaviour trees. Here we present the number of nodes in the automatically constructed explanation model, the number of counterfactual candidates (i.e. interventions) considered during the counterfactual search, and the number of explanations in the explanation set.}
\label{tab:random_results_numbers}
\resizebox{\textwidth}{!}{%
\begin{tabular}{cc|ccc|ccc|ccc}
\multicolumn{2}{c|}{\textbf{BT/State Parameters}} & \multicolumn{3}{c|}{\textbf{Number of Explanation Model Nodes}} & \multicolumn{3}{c|}{\textbf{Number of Counterfactual Candidates}} & \multicolumn{3}{c}{\textbf{Number of Explanations}} \\
\textbf{No. State Vars}      & \textbf{No. BT Leaves}     & \textbf{Min.}          & \textbf{Max.}          & \textbf{Mean (Std)}            & \textbf{Min.}           & \textbf{Max.}          & \textbf{Mean (Std)}             & \textbf{Min.}       & \textbf{Max.}       & \textbf{Mean (Std)}       \\ \hline
4                   & 2                 & 10            & 14            & 11.71 (1.19)          & 3              & 14            & 6.51 (3.61)            & 1          & 4          & 2.12 (0.95)      \\
4                   & 4                 & 19            & 29            & 22.52 (2.95)          & 3              & 25            & 6.79 (6.08)            & 1          & 5          & 2.10 (1.24)      \\
4                   & 8                 & 33            & 51            & 41.36 (5.21)          & 4              & 52            & 19.88 (15.90)          & 1          & 6          & 2.95 (1.31)      \\
8                   & 2                 & 14            & 24            & 16.83 (2.14)          & 3              & 23            & 8.01 (4.66)            & 1          & 7          & 3.04 (1.35)      \\
8                   & 4                 & 22            & 45            & 29.04 (5.47)          & 3              & 43            & 10.67 (8.90)           & 1          & 13         & 3.34 (2.42)      \\
8                   & 8                 & 43            & 77            & 57.73 (9.06)          & 5              & 81            & 22.47 (21.57)          & 1          & 13         & 3.98 (2.83)      \\
12                  & 2                 & 18            & 33            & 23.68 (4.48)          & 3              & 31            & 11.86 (6.74)           & 1          & 11         & 3.77 (2.02)      \\
12                  & 4                 & 28            & 60            & 40.82 (9.98)          & 3              & 61            & 17.23 (14.64)          & 1          & 12         & 3.83 (2.51)      \\
12                  & 8                 & 50            & 105           & 77.90 (14.19)         & 6              & 88            & 34.04 (21.26)          & 1          & 14         & 5.70 (3.08)      \\ \hline
All                 & All               & 10            & 105           & 39.96 (23.13)         & 3              & 88            & 17.02 (17.03)          & 1          & 14         & 3.71 (2.52)     
\end{tabular}}
\end{table*}

The majority of executions (1113, $\sim$61.8\%) result in no difference whatsoever in the sequence of node executions, return statuses and decisions made (see Table \ref{tab:random_results_target_runtime}), which can be attributed to the particular implementation of the BT functions and the arbitrary selection of default initialisations. Nevertheless, in a large number of executions (687, $\sim$38.2), the perturbations result in different behaviour between executions, for which a valid query regarding the difference could be automatically constructed. Across all of these executions our approach achieves 100\% \textit{TRR}, meaning the explanation set generated by Algorithm \ref{alg:counterfactual_search} always contains the target variable. This indicates that our algorithm is capable of producing explanations that correctly identify the cause of a difference in behaviour across a wide class of BTs and state graphs.

Examining the runtime for each assignment of BT and state parameters (recorded as the amount of time taken to generate the explanation set upon receiving a query), we see that in all cases our approach generates explanations very quickly, never exceeding half a second for even the largest BTs and states. Given the time it may take a user to produce a query and subsequently take in and process one or more explanations, such runtimes are certainly suitable for real-time robotics applications.

Regarding the number of nodes in the explanation model, we see a general trend in which the number of nodes grows with the number of BT nodes and the number of state variables, as predicted in Section \ref{sec:complexity_explanation_model}. The variability present in Table \ref{tab:random_results_numbers} is due to the different seeds used for each parameter set and the different degrees of connectivity.\looseness=-1

Similarly, the number of counterfactual candidates considered during the counterfactual search (Algorithm \ref{alg:counterfactual_search}) also grows with the size of the BT and/or state, due to the larger explanation model over which the search is conducted. Once again, we observe high variability in the number of interventions considered, this time due to the position of the queried node in the explanation model. If the query targets a node with few ancestors, then much of the explanation model is discarded in the search, and thus even a very large explanation model could result in a small counterfactual search. The number of ancestors a node has in the explanation model depends on how far right the corresponding BT node is in the tree, as well as the ``connectivity'' and dynamics of the state.

Finally, we observe that the number of explanations in the explanation set also grows with the number of state variables, due to the larger number of counterfactual candidates which are considered. We see that the number of BT nodes does not seem to have much of an effect on the size of the explanation set, which we hypothesise is due to the \textit{target recovery} metric itself, which explains the first difference between executions, which is more likely to occur earlier in the BT. We also note that, while the number of explanations returned is usually quite small (mean of 5.70 for the largest problem), outlier explanation sets can be quite large (maximum of 14 for the largest problem). In these cases, it is clear that further search filters, such as those discussed in Section \ref{sec:architecture_properties}, are required to produce manageable explanation sets. Ultimately, the size of the explanation set is heavily dependent on the size of the problem (i.e. the size of the explanation model), the variable queried, and the particular implementation of the functions in $\mathcal{F}$.

\subsection{Example: Serial Recall Task}\label{sec:evaluation_serial_recall}
While the experiments in the previous section show the applicability of our method to a wide class of BT and state graph structures, we also demonstrate the same \textit{target recovery} metric in a larger, more complex problem that more closely matches a real-world application. In this case, we model the serial recall use case. Serial recall is the ability to recall a novel sequence in order, and can be measured using memory tasks where participants are provided a sequence of ``items'' (e.g. verbally, visually, with physical objects, etc.) and must correctly recall the order of the sequence \citep{hurlstone2024serial}. We model such a task where the sequence is provided by a social robot. The use of social robots in cognitive exercise and therapy is steadily emerging~\citep{yuan2021systematic,andriella2022introducing,cavallaro2024social}, and the approach we outline here seeks to emulate the complexity of the decision-making in these real world applications in simulation.\looseness=-1

In our formulation, the robot can select the length of the sequence (i.e. the number of items) and the complexity (number of unique items). Once the user repeats back the sequence, it can evaluate the sequence and, depending on the response time and accuracy in repeating the sequence, can choose to continue to a new sequence, repeat the existing sequence, or end the entire exercise. In between showing the sequences, the robot can also select ``social actions'' such as encouraging or congratulating the user, or trying to recapture their attention. The task repeats until a maximum number of sequences is reached or the robot decides to end the task prematurely.\looseness=-1

Apart from task state variables set by the system, the state also consists of a model of the user. Adapting the ``persona'' of \cite{andriella2019learning}, we define a user along several dimensions, namely Attention, Memory, Reactivity and Frustration, each ranging from 0 (low) to 1 (high). These factors, together with the difficulty of the task, in turn affect other user variables such as Confusion and Engagement, as well as the response time and accuracy of the user when repeating a sequence.

We use a BT to implement the robot's decision-making, and have it interact with a simulated user. We also construct a state model to represent causal assumptions about the user which correspond to rules in the simulation (e.g. such as the influence of user attention and confusion on engagement, or the influence of reactivity and sequence length and complexity on the user's response time). Some noise is introduced to the user's accuracy and response time to allow for differences between executions. The noise is controlled by a predetermined seed so as to not violate the assumptions that the behaviour and state are deterministic (see Section \ref{sec:extending_bts}). After fully implementing the simulated use case, the resulting BT has 33 nodes (of which 18 are leaf nodes), and the state is represented as a configuration of 26 state variables, some discrete and some continuous (both are outlined more extensively in Appendix \ref{app:serial_recall_task}). The resulting explanation model built by Algorithm \ref{alg:causal_graph} contains 177 nodes and 406 edges, demonstrating the complexity of this task.\looseness=-1

To evaluate our approach on this use case, we again employ the \textit{target recovery} metric. This time, we define a ``default'' user profile with initial values representing a typical user (setting Attention, Reactivity and Memory to 0.8 and Frustration to 0). The altered initialisations are generated by constructing alternate user profiles with a single perturbed user attribute (\textit{Frustrated} sets Frustration to 1, \textit{No Attention}, \textit{No Reactivity} and \textit{No Memory} set Attention, Reactivity and Memory to 0, respectively). Each profile is executed across 50 different seeds, and thus there are 50 comparisons between the default profile and each altered profile. The results of the \textit{target recovery} metric, together with runtime statistics, the number of counterfactual candidates considered and the number of explanations in the explanation set, are provided in Table \ref{tab:cog_seq_results}.

\begin{table*}[]
\caption{Results for target recovery experiments in the serial recall use case. The behaviour tree has 33 nodes (18 leaves) and uses a state consisting of 26 variables. The resulting explanation model contains 177 nodes and 406 edges.}
\label{tab:cog_seq_results}
\resizebox{\textwidth}{!}{%
\begin{tabular}{c|cc|c|ccc|ccc|ccc}
\multirow{2}{*}{\textbf{Profile}} & \multicolumn{2}{c|}{\textbf{Number of Runs}}    & \multirow{2}{*}{\textbf{TRR}} & \multicolumn{3}{c|}{\textbf{Runtime (s)}}            & \multicolumn{3}{c|}{\textbf{No. Counterfactual Candidates}} & \multicolumn{3}{c}{\textbf{No. Explanations}} \\
                                  & \textbf{Target Found} & \textbf{No Difference} &                                                & \textbf{Min.} & \textbf{Max.} & \textbf{Mean (Std)} & \textbf{Min.}      & \textbf{Max.}     & \textbf{Mean (Std)}     & \textbf{Min.} & \textbf{Max.} & \textbf{Mean (Std)} \\ \hline
Frustrated                        & 47                    & 3                      & 1                                              & 0.6752        & 1.3029        & 0.7122 (0.0891)     & 396                & 545               & 399.49 (21.69)          & 1             & 7             & 1.13 (0.88)         \\
No Attention                      & 50                    & 0                      & 1                                              & 0.0413        & 0.884         & 0.5464 (0.4033)     & 122                & 449               & 321.58 (159.37)         & 5             & 21            & 11.22 (4.35)        \\
No Reactivity                     & 49                    & 1                      & 1                                              & 0.041         & 0.3973        & 0.0806 (0.1070)     & 122                & 294               & 139.55 (52.60)          & 3             & 37            & 7.31 (9.89)         \\
No Memory                         & 50                    & 0                      & 1                                              & 0.6821        & 0.8961        & 0.7446 (0.0694)     & 396                & 449               & 406.60 (21.42)          & 4             & 37            & 18.64 (13.43)       \\ \hline
All                               & 196                   & 4                      & 1                                              & 0.041         & 1.3029        & 0.5203 (0.3425)     & 122                & 545               & 316.44 (137.50)         & 1             & 37            & 9.71 (10.70)       
\end{tabular}}
\end{table*}

In 196 (98\%) of the comparisons, the perturbation resulted in a different in behaviour between the default and the altered profiles. And in every single case, our approach successfully recovered the target variable among the explanation set, demonstrating its correctness even for a more complex BT and state model. Despite the much larger size and complexity of the serial recall BT when compared to the randomly generated ones in Section \ref{sec:evaluation_random}, the time taken to generate explanations remains low enough for applicability in real-time robotics use cases.

Examining the number of counterfactual candidates and number of explanations returned for each altered profile, which are larger for this problem than the smaller randomly generated BTs as presented in Table \ref{tab:random_results_numbers}, we see the dependence of the sizes of the search and explanation set on not only the size of the explanation model but also the part of the model that is queried. For example, the changes triggered by the \textit{Frustrated} profile result in queries that produce relatively small explanation sets (mean 1.13), despite a relatively large number of counterfactual candidates considered. This is in part because the effects of the \textit{Frustration} variable are limited, with only a single behaviour node having the variable in its input set (though one that occurs towards the right-most edge of the BT, hence the large number of counterfactual candidates). The changes triggered by the \textit{No Memory} profile, on the other hand, result in much larger explanation sets (mean 18.64) and a greater search space (mean 406.60), with the \textit{Memory} variable and its descendants impacting more nodes in the BT (5 nodes). The variance between explanation set sizes can also be explained by the difference in the implementations of each node (i.e. the functions in $\mathcal{F}$). Overall, we observe that a number of factors, including the size of the BT and state, the nature of the functions in $\mathcal{F}$, and the variable being queried, have the possibility of producing overly large sets of explanations in certain cases. If we are to reduce this set of explanations to a smaller, more relevant set, this notion of relevance must be encoded for, for example, by applying filters to the counterfactual search that account for properties such as discriminative power or actionability, as discussed in Section \ref{sec:limitations}.

\subsection{Comparison with an LLM Baseline}\label{sec:llm_baseline}
While most of the methods discussed in Section \ref{sec:related_work} cannot be directly compared to our method as they either address different types of queries (e.g. answering ``what'' questions instead of ``why'' questions) or can only provide a limited subset of causal explanations (as discussed in Section \ref{sec:comparison_with_literature}), approaches that leverage the ``common sense'' of LLMs are able to provide explanations in response to any natural language query and are thus suitable for comparison. Indeed, as discussed in Section \ref{sec:related_work}, several works address the problem of LLM-based explanations for robots~\citep{frering2025integrating,gebelli2025personalised}, including the work of \cite{tagliamonte2024generalizable} which specifically targets behaviour trees. In this section, we examine the results of some variations of an LLM-based approach inspired by \cite{tagliamonte2024generalizable} in the serial recall domain, and compare against the results obtained in Section \ref{sec:evaluation_serial_recall}.\looseness=-1

To make the comparison as fair as possible, we give the LLM access to all the same information that our method uses to generate explanations. This includes textual descriptions of the task, the environment state (including all information contained in the state model), and BT (including its structure, input set, output set, action set, and detailed behavioural description), as well as the raw episodic memory. This information is fed to the LLM in the form of a system prompt\footnote{The full prompts are available at \prompturl}, which also details the expected format of the explanation and examples of correct explanations (generated by our approach on different seeds than those tested in the evaluation).

We compare two versions of the prompt: a \textit{simple} case, in which the LLM is instructed to provide only a single reason of the form $Var=X$, and a \textit{complete} case, in which the LLM is instructed to provide multiple explanations of the form $\langle \mathbf{R}, (\mathbf{J},\mathbf{K}) \rangle$, as defined in Section \ref{sec:definitions_counterfactuals}. For each version of the prompt, we compare two models: \textit{phi4}~\citep{abdin2024phi} and \textit{deepseek-r1:32b}~\citep{guo2025deepseek}, with 14B and 32B parameters respectively. Both models were run locally using Ollama\footnote{\url{https://docs.ollama.com/}}, so that runtimes could be compared with those reported in the previous sections.

In addition to the \textit{target recovery rate} (TRR) and runtime measured in previous sections, we also measure the rate at which the LLM identifies variables which are in the explanation set produced by our method (\textit{correct reason variable}), the rate at which these correctly identified variables have the correct value according to the episodic memory (\textit{correct reason value}), and the number of runs in which the LLM produced reasons using variables which do not exist in either the state model or the BT (\textit{incorrect variable}). We also note the number of runs in which a formatting error renders the explanation unusable by the evaluation system. Results are collected over 200 runs (50 per profile), using the same seeds and profiles as in Section \ref{sec:evaluation_serial_recall}, and are tabulated in Table \ref{tab:llm_results}.

\begin{table}[]
\caption{Results for evaluations performed in the serial recall domain with two LLMs with two sets of prompts. We display the number of runs in which explanations were successfully generated (\textit{Explained}), the number of runs in which no difference in execution occurs (\textit{No Difference}), and the number of runs in which a formatting error creates an unusable explanation (\textit{Error}). We also display the \textit{target recovery rate} (\textit{TRR}), the ratio of explanations in which the identified reason variable is a viable causal explanation (\textit{Variable}), the ratio of explanations which also correctly attributed the variable value from the episodic memory (\textit{Value}), the number of runs in which a variable not present in the state model or BT is used in an explanation (\textit{Incorrect}), and runtime statistics. Evaluations are performed on the same hardware, seeds and profiles as in Section \ref{sec:evaluation_serial_recall}, and thus results are directly comparable to those presented in Table \ref{tab:cog_seq_results}.}
\label{tab:llm_results}
\resizebox{\textwidth}{!}{%
\begin{tabular}{lc|ccc|c|cc|c|ccc}
\multicolumn{1}{l}{\multirow{2}{*}{\textbf{Model}}} & \multirow{2}{*}{\textbf{Prompt}} & \multicolumn{3}{c|}{\textbf{Number of Runs}}                         & \multirow{2}{*}{\textbf{TRR}} & \multicolumn{2}{c|}{\textbf{Correct Reason}} & \textbf{Incorrect} & \multicolumn{3}{c}{\textbf{Runtime (s)}}            \\
\multicolumn{1}{c}{}                                &                                  & \textbf{Explained} & \textbf{No Difference} & \textbf{Error} &                               & \textbf{Variable}      & \textbf{Value}     &  \textbf{Variable} & \textbf{Min.} & \textbf{Max.} & \textbf{Mean (Std)} \\ \hline
\multirow{2}{*}{phi4}                               & Simple                           & 190                & 4                      & 6                     & 0.06 (0.24)                   & 0.04 (0.15)            & 0.22 (0.33)        & 188                                          & 0.9586        & 4.3452        & 2.3477 (0.5874)     \\
                                                    & Complete                         & 195                & 4                      & 1                     & 0.20 (0.40)                   & 0.22 (0.36)            & 0.35 (0.40)        & 161                                          & 2.356         & 13.1847       & 4.3336 (1.2712)     \\ \hline
\multirow{2}{*}{deepseek-r1:32b}                    & Simple                           & 195                & 4                      & 1                     & 0.45 (0.50)                   & 0.59 (0.48)            & 0.58 (0.49)        & 8                                            & 4.9088        & 95.2549       & 14.3402 (9.9605)    \\
                                                    & Complete                         & 179                & 4                      & 17                    & 0.53 (0.50)                   & 0.53 (0.31)            & 0.21 (0.28)        & 31                                           & 11.9767       & 65.0887       & 23.8133 (6.8669)   
\end{tabular}}
\end{table}

First of all, we note that the \textit{target recovery rate} is low, even for the larger model. When compared to the 100\% \textit{target recovery rate} achieved by our method, the necessity for explicit causal modelling of the task becomes clear. Even though the LLMs have access to all the necessary information through the prompt, they lack the ability to causally reason and chain together causes and effects in a consistent way, which is essential for providing causal explanations, especially for temporal sequences of events where cause and effect may be separated in time. We also note that the larger model unsurprisingly performs significantly better, which holds for the remaining metrics (aside from runtime). We also note that the \textit{complete} prompt results in improved \textit{target recovery rate}, likely due to the LLM producing more explanations, which gives a higher probability of selecting the correct target variable.

Setting aside the \textit{target recovery metric} and considering the complete possible set of correct explanations, we observe similar rates at which the LLMs identify correct reason variables in their explanation sets (\textit{Correct Reason - Variable} in Table \ref{tab:llm_results}). These results indicate that the models, especially the smaller model, consistently identify reasons which are not causally correct (i.e. intervening on that variable would not result in satisfying the query when conducting the counterfactual search in Algorithm \ref{alg:counterfactual_search}). Furthermore, when the LLMs do identify a correct reason variable, they very often attribute incorrect values in the reason (\textit{Correct Reason - Variable} in Table \ref{tab:llm_results}), likely due to the difficulty in retrieving the appropriate values from a potentially very large episodic memory. Indeed, the generative approach taken by LLMs often results in so-called ``hallucinations'', where responses are either incorrect or nonsensical \citep{huang2025survey}. This effect is most clear when considering the number of runs for which the LLMs referenced variables that did not exist in either the state model or the BT (\textit{Incorrect Variable} in Table \ref{tab:llm_results}). This is especially a problem for the smaller model, which almost always included at least one ``hallucinated'' reason in the explanation set. In many cases, these incorrect variables may clearly reference a real variable (e.g. ``UserEngagementLevel'', while incorrect, is very similar to the real variable ``UserEngagement''), some are completely untethered to the state and BT information provided in the prompt (e.g. ``CompletionStatus'' or ``SequenceEffectiveness''). Both models also occasionally produced explanations which disregarded formatting instructions, rendering the evaluation system unable to extract the reasons from the model output. The inconsistency in output formatting poses challenges to integrating an LLM-based explanation module into a larger architecture, and together with the problem of ``hallucination'', casts further doubt on the applicability of LLMs in use cases where correct explanations are of special importance. Our method, on the other hand, consistently generates explanations which are grounded in the state model and BT structure and are always consistently formatted by design, making integration into larger architectures considerably simpler.

Finally, we note that the LLMs take orders of magnitude longer to produce explanations. In the best case (\textit{phi4 - Simple}), the model runs on average 4.5 times slower than our method, while in the worst case (\textit{deepseek-r1:32b - Complete}), the average runtime is 45.8 times slower. LLM explanation runtimes, in addition to being considerably slower than our method (the minimum runtime in Table \ref{tab:llm_results} is within half a second of the maximum runtime of our method in Table \ref{tab:cog_seq_results}), also suffer from high variance, rendering these models impractical for real-time integration into robotics systems.\looseness=-1

To conclude this comparison, it is clear that an LLM-based approach is vastly inferior to our own method in terms of both explanation correctness and runtime, highlighting the need for causal reasoning in explanation generation.

%%%
%%% NEW: LIMITATIONS AND FUTURE WORK
%%%

\section{Limitations and Future Work}\label{sec:limitations}

While our proposed architecture for generating causal explanations from behaviour tree executions is capable of producing accurate explanations of decisions, executions, node returns and state values, expanding the scope of causal explanations for BTs from the related work identified in Section \ref{sec:related_work}, it nevertheless carries limitations which should be addressed in future research.

Likely the biggest limitation of our approach is the assumption that the state and BT are completely deterministic, as discussed in Section \ref{sec:extending_bts}. Such an assumption is required to ensure that explanations are consistent and correct, as the counterfactual search directly calls the functions in $\mathcal{F}$. Relaxing this assumption is an important next step for this research, and may require treating the values of nodes in the explanation model as probability distributions rather than deterministic function outputs, such as the causal Bayesian networks used by \cite{diehl2022did}. In doing so, the extended architecture could account for uncertainty in both the inputs to explanation model functions (such as uncertain states caused by unobserved state variables or noisy sensors) and the outputs of said functions, allowing the architecture to handle stochastic functions within the behaviour tree and uncertain outcomes of decisions (e.g. due to failures in robot action executions).

Another important consideration is the need to inject domain knowledge in the form of a state model into the architecture in order to produce the final explanation model (see Section \ref{sec:architecture_causal_model_domain_knowledge}). Indeed, as demonstrated in Section \ref{sec:case_study_domain_knowledge}, such domain knowledge is key to allowing the explanation model to recover certain target variables in its explanations. In this work, we remain agnostic to the source of the domain knowledge, focusing instead on its utility in constructing the explanation model. However, in the service of fully automating the explanation model construction, an important component of future work may be to investigate methods of automating the construction of the state model. As the state model encodes causal relations, its automatic construction requires causal discovery, which is an active research area that has produced a variety of methods~\citep{wang2024survey}. Promising directions may be to extract the causal structure and functions of the state model from knowledge structures such as ontologies~\citep{ben2009integrating,stegnar2024enhancing}, or to use learning-based approaches from real-world data or simulation~\citep{castri2023enhancing,edstrom2023robot}. 

Furthermore, as discussed in Section \ref{sec:complexity_counterfactual_search} and as can be seen in Tables \ref{tab:random_results_target_runtime}, \ref{tab:random_results_numbers} and \ref{tab:cog_seq_results},  the size of the explanation set returned by the counterfactual search can vary significantly depending on the size and complexity of the explanation model, as determined by the BT and state model. While our approach guarantees that all valid, minimal counterfactuals are found, it may be useful to restrict this search to provide a smaller, more relevant set of explanations, as is generally preferred by users \citep{miller2019explanation}. This may be achieved by adding further restrictions during the counterfactual search, based on desirable properties such as \textit{discriminative power} (as indicated by the critical influences discussed by \cite{albini2020relation}) or \textit{plausibility} (which could be based on data regarding the most probable state configurations). Further relevance criteria could be devised, such as using the distance between nodes in the explanation model, or by further creating stricter queries using contextual information and models of the user. Applying criteria to reduce the search space also has the effect of reducing the complexity of the counterfactual search, which, as noted in Section \ref{sec:complexity_counterfactual_search}, is greatly impacted by the size of the explanation model. Additionally, LLMs represent a promising tool which could see applications in reducing or summarising an explanation set after the counterfactual search has been conducted~\citep{dhaini2024explainability}. 

In this work, our evaluation focuses on functionally-grounded approaches, excluding human- or application-grounded evaluations which require user studies conducted with human participants~\citep{doshi2017towards}. While user studies are necessary in order to evaluate the effect of explanations on desiderata such as understandability and trust, we argue it is crucial to validate explanation generation functionally, in terms of properties such as correctness and runtime efficiency, before testing with users. Having performed such validations in this work, evaluating our approach in user studies becomes possible. Such user studies require explanation generation to be embedded into a larger human-robot interaction, requiring interfaces for providing queries and for communicating explanations~\citep{anjomshoae2019explainable,matarese2021user}. Such interfaces require additional design decisions than the ones we have taken in this work to generate the content of explanations. Our architecture requires formal representations of queries and produces formal representations of explanations (as defined in Section \ref{sec:definitions_counterfactuals}). Leveraging LLMs to convert natural language questions into formal explanation queries, sensitive to question context within an environment and a broader human-robot interaction, may prove to be fruitful. Similarly, LLMs may be used to convert formal explanations or simple templates into natural language explanations, whose effectiveness in communicating information can be evaluated in studies with human participants. Aside from natural language, other modalities such as visual interfaces (e.g. annotated BTs or causal models) can be explored~\citep{hoque2021outcome}.

In our comparison with an LLM-based baseline in Section \ref{sec:llm_baseline}, we demonstrated that the LLMs produced, on average, less accurate explanations with much longer runtimes than our approach. One limitation of this comparison is that our baseline likely does not represent the best performance an LLM-based approach could achieve for explainability. In particular, our approach is end-to-end and monolithic, producing explanations with a single, large prompt. Better performance may be possible with a more ``agentic'' approach, in which functionality of the method is divided into a chain of LLM agents performing specific tasks (e.g. interpretting a query, identifying relevant parts of the episodic memory, chaining together causes and effects, synthesising an explanation set, etc.). However, we hypothesise that the problems identified with our LLM baseline, namely target recovery, the inclusion of incorrect information, and poor runtime performance, are likely to persist in such an approach, due to the ``inevitability'' of ``hallucinations''~\citep{xu2024hallucination}. Of course, this does not preclude LLMs from being useful components of an explainability system. Indeed, areas which are not addressed by our method may be more suitable to LLMs, as discussed above. However, it is clear that the process of explanation generation, at least in terms of identifying reasons for BT behaviour and other events, is better left to our method, which is produces consistent and correct explanations in a practical time frame.\looseness=-1

Finally, further work could also be done to expand the range of decision-making structures from which an explanation model can be built. The class of BTs can be expanded by including other important node types, such as parallel nodes and decorators. Other common types of decision-making structures can also be explored, such as finite state machines \citep{iovino2024comparison} and planning frameworks such as hierarchical task networks \citep{hayes2016autonomously}.\looseness=-1

%%%
%%% CONCLUSION
%%%

\section{Conclusion}\label{sec:conclusion}
In conclusion, we have presented a novel approach to generating counterfactual, causal explanations from BTs, leveraging a causal model automatically built from the BT structure and domain knowledge of the state and individual BT nodes. Our method is able to answer contrastive queries about node executions and return statuses, decisions made by a BT and the value of state variables in the environment. As the explanation model is built from the BT structure, state model and BT node inputs and outputs, our method is able to use all of these sources of information when generating explanations, and can take into account events earlier in the execution when producing explanations about later events. When compared to other methods, our method differs from model-agnostic approaches in its ability to leverage BT structure and domain knowledge to generate causal explanations. It differs from other BT-based approaches which do take advantage of BT structure in that it is able to answer a wide range of contrastive questions with reasons beyond a trace of executions. Finally, it differs from generative (i.e. LLM-based) approaches which are able to answer these types of questions in that the explanations are guaranteed to respect the causality of the system, with fidelity proportional to that of the injected domain knowledge.

Through evaluations in both a wide range of randomly generated problems and with a larger BT simulating a real HRI use case, we have demonstrated that the explanations generated by our approach are correct and can accurately recover the causes of differences in behaviour. Moreover, our evaluations have shown that these explanations can be generated in real time, suitable for robotics applications. This stands in contrast to LLM-based approaches which are not only more computationally demanding but are also inconsistent in their accuracy.

As discussed in Section \ref{sec:limitations}, there are a number of lines of potential future that are open which may address the limitations of this work, particularly in the automatic learning of relevant domain knowledge, probabilistic approaches for handling stochasticity and uncertainty, and leveraging notions of actionability, plausability and relevance for reducing the size of the counterfactual search space and/or explanation set. Further work can also be done to validate approaches built from this framework in user studies with real environments and physical robots, requiring the appropriate interfaces to be built for query formulation and explanation communication. Ultimately, these extensions require a solid foundation of accurate, causal explanations which are provided by our method.

In summary, not only is our method a robust approach to generating correct counterfactual explanations for BTs, but it also sets the foundation for the future development of transparent, interpretable robots capable of interactively and automatically explaining their decisions, which in turn has the potential to improve HRI by fostering increased understandability and trust.\looseness=-1


% TODO: Anonymise
\section*{Acknowledgements}
This work was supported by Horizon Europe under the MSCA grant agreement No 101072488 (TRAIL).

\appendix
\section{Algorithms}\label{app:algorithms}
In this appendix we expand on the algorithms presented in Section \ref{sec:architecture} with more detailed pseudocode. Algorithm \ref{alg:cm_from_bt_full} depicts the complete pseudocode of Algorithm \ref{alg:cm_from_bt}, as described in Section \ref{sec:architecture_causal_model_bt_structure}. Algorithm \ref{alg:add_temporal_edges} depicts the temporal linking function used in Algorithm \ref{alg:cm_from_dk}, described in Section \ref{sec:architecture_causal_model_domain_knowledge}.  Algorithm \ref{alg:functions}, which in turn makes use of Algorithms \ref{alg:execution} - \ref{alg:state_node}, describes the functions $\mathcal{F}$ of the nodes in the explanation graph $\mathcal{G}$, as described in Section \ref{sec:architecture_causal_model_functions}.

\begin{figure}[t]
    \centering
    \includegraphics[width=\linewidth]{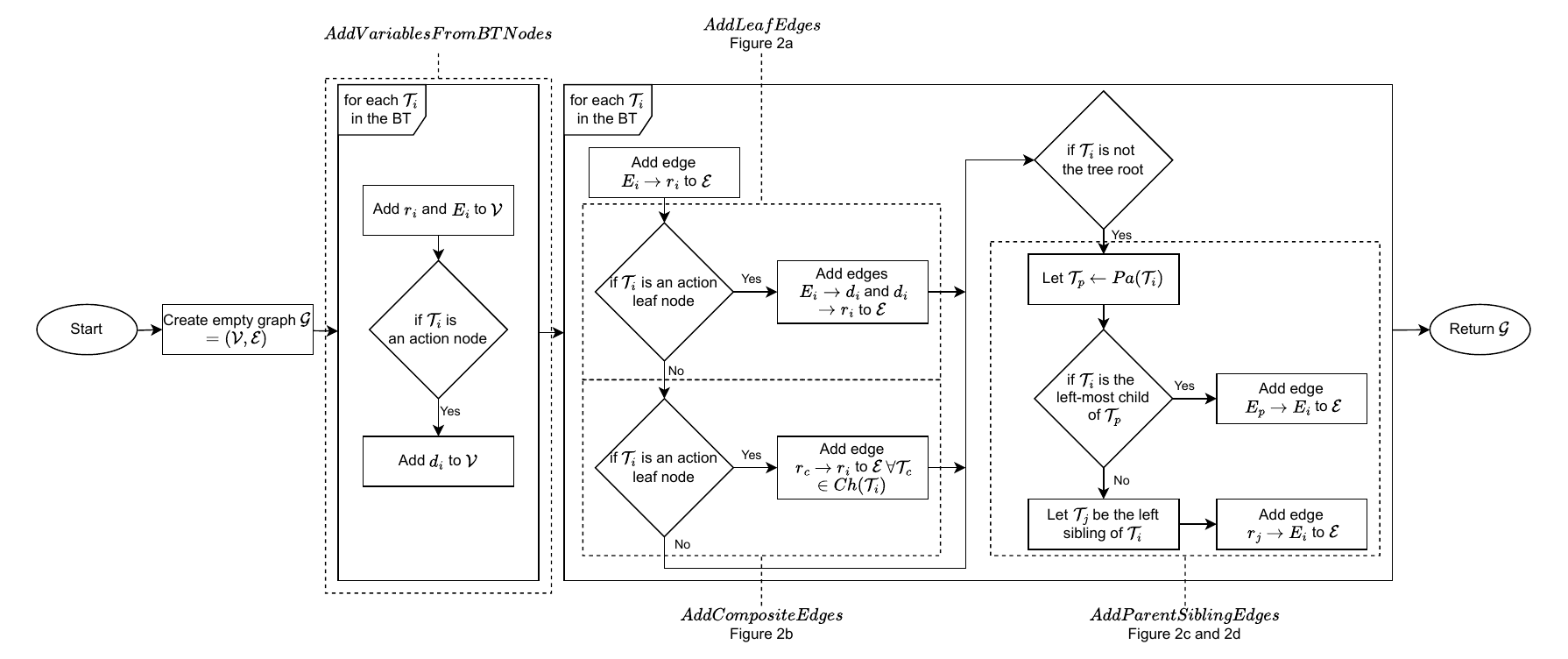}
    \caption{\footnotesize
    A flowchart depicting the process of constructing the explanation model from the BT structure and state model, as described in Section \ref{sec:architecture_causal_model}. In particular, the figure represents the procedure undertaken by Algorithm \ref{alg:causal_graph}.}
    \label{fig:flowchart_build_model}
\end{figure}

\begin{figure}[t]
    \centering
    \includegraphics[width=\linewidth]{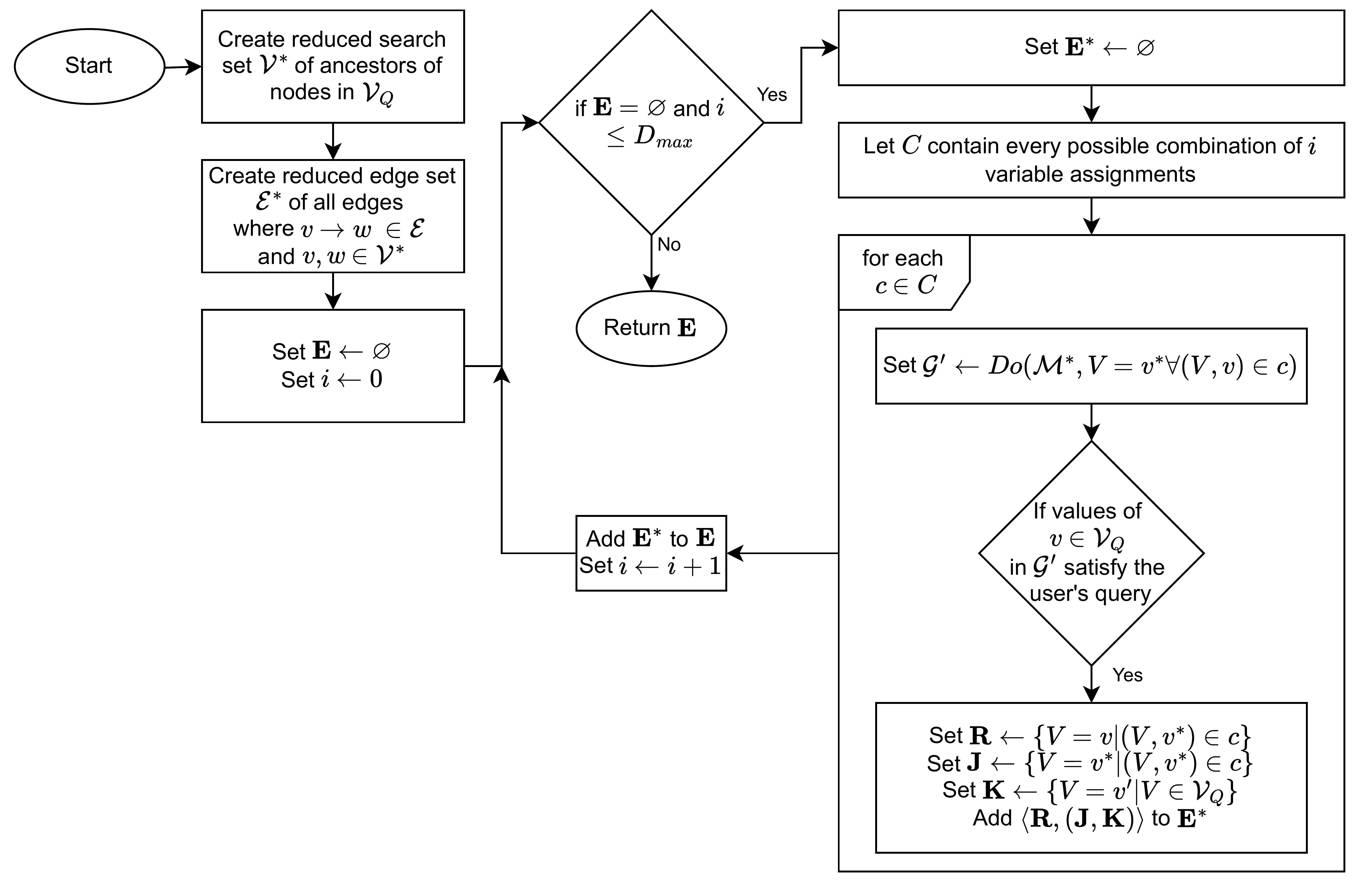}
    \caption{\footnotesize
    A flowchart depicting the counterfactual search, which performs interventions on the explanation model in order to build a set of causal explanations, as described in Section \ref{sec:architecture_explanation_generation} and presented in Algorithm \ref{alg:counterfactual_search}.}
    \label{fig:flowchart_counterfactual_search}
\end{figure}

In addition to pseudocode, we also present two flowcharts which provide a visual depiction of the key contributions of this work. The procedure for building the explanation model is depicted in Figure \ref{fig:flowchart_build_model}, while the procedure for performing the counterfactual search is depicted in Figure \ref{fig:flowchart_counterfactual_search}.

\begin{algorithm}
    \caption{Constructing the explanation graph from the behaviour tree structure - complete pseudocode}\label{alg:cm_from_bt_full}
    \begin{algorithmic}[1]
    \Function{GraphFromStructure}{ } \funclabel{algf:graph_from_structure_full} \label{algl:graph_from_structure_full}
        \State $\mathcal{G} \gets \varnothing$
        \State $\mathcal{G} \gets$ \Call{AddVariablesFromBTNodes}{$\mathcal{G}$}
        \For {node $\mathcal{T}_i$ in the behaviour tree}
            \State Add edge $E_i \to r_i$ to $\mathcal{E}$\label{algl:add_E_to_R_edge_full}
            \If{$\mathcal{T}_i$ is a leaf node}
                \State $\mathcal{G} \gets$ \Call{AddLeafEdges}{$\mathcal{G}$, $\mathcal{T}_i$}
            \Else
                \State $\mathcal{G} \gets$ \Call{AddCompositeEdges}{$\mathcal{G}$, $\mathcal{T}_i$}
            \EndIf
            \If{$\mathcal{T}_i$ is not the tree root}
                \State $\mathcal{G} \gets$ \Call{AddParentSiblingEdges}{$\mathcal{G}$, $\mathcal{T}_i$}
            \EndIf
        \EndFor
        \State \textbf{return} $\mathcal{G}$
    \EndFunction

    \Function{AddVariablesFromBTNodes}{$\mathcal{G}$}\funclabel{algf:add_variables_from_bt_nodes_full} \label{algl:add_variables_from_bt_nodes_full}
        \For {node $\mathcal{T}_i$ in the behaviour tree}
            \State Add variables $r_i$ and $E_i$ to $\mathcal{V}$
            \If{$\mathcal{T}_i$ is an action node}
                \State Add variable $d_i$ to $\mathcal{V}$
            \EndIf
        \EndFor
        \State \textbf{return} $\mathcal{G}$
    \EndFunction

    \Function{AddLeafEdges}{$\mathcal{G}$, $L_i$}\funclabel{algf:add_leaf_edges_full} \label{algl:add_leaf_edges_full}
        \If{$L_i$ is an action node}
            \State Add edge $E_i \to d_i$ to $\mathcal{E}$
            \State Add edge $d_i \to r_i$ to $\mathcal{E}$
        \EndIf
        \State \textbf{return} $\mathcal{G}$ \label{algl:add_leaf_edges_end_full}
    \EndFunction

    \Function{AddCompositeEdges}{$\mathcal{G}$, $\mathcal{T}_i$}\funclabel{algf:add_composite_edges_full} \label{algl:add_composite_edges_full}
        \For{each child $\mathcal{T}_c \in Ch(\mathcal{T}_i)$}
            \State Add edge $r_c \to r_i$ to $\mathcal{E}$
        \EndFor
        \State \textbf{return} $\mathcal{G}$ \label{algl:add_composite_edges_end_full}
    \EndFunction

    \Function{AddParentSiblingEdges}{$\mathcal{G}$, $\mathcal{T}_i$}\funclabel{algf:add_parent_sibling_edges_full} \label{algl:add_parent_sibling_edges_full}
        \State $\mathcal{T}_p \gets Pa(\mathcal{T}_i)$
        \If{$\mathcal{T}_i$ is the left-most child of $\mathcal{T}_p$}
            \State Add edge $E_{p} \to E_i$ to $\mathcal{E}$
        \Else
            \State Let $\mathcal{T}_j$ be the left sibling of $\mathcal{T}_i$
            \State Add edge $r_j \to E_i$ to $\mathcal{E}$
        \EndIf
        \State \textbf{return} $\mathcal{G}$ \label{algl:add_parent_sibling_edges_end_full}
    \EndFunction
    \end{algorithmic}
\end{algorithm}

\begin{algorithm}
    \caption{Linking temporal versions of state variables}\label{alg:add_temporal_edges}
    \begin{algorithmic}[1]
        \Function{AddTemporalEdges}{$\mathcal{G}$,$\tau$} \funclabel{algf:add_temporal_edges} \label{algl:add_temporal_edges}
        \For{$X \in \mathcal{X}$}
            \If{$X$ has no parents in $\mathcal{G}_\mathcal{S}$}
                \For{$t \in [0,\tau(X))$}
                    \State Add edge $X^{(t)} \to X^{(t+1)}$ to $\mathcal{E}$
                \EndFor
            \EndIf
        \EndFor
        \State \textbf{return} $\mathcal{G}$\label{algl:add_temporal_edges_end}
    \EndFunction
    \end{algorithmic}
\end{algorithm}

\begin{algorithm}
    \caption{Defining $\mathcal{F}$ for each node in the explanation model}\label{alg:functions}
    \begin{algorithmic}[1]
    \Function{$\mathcal{F}_n$}{$\pmb{v}$} \funclabel{algf:function} \label{algl:function}
        \If{$n \corresponds E_i$ for some node $\mathcal{T}_i$}
            \State \textbf{return} \Call{Execution}{$\mathcal{T}_i$}
        \EndIf
        \State Let $\mathcal{P} = \{p|p\text{ is a parent of }n\text{ in }\mathcal{G}\}$
        \State $\pmb{s} \gets $\Call{SetInputState}{$n,\pmb{v},\mathcal{P}$}
        \If{$n \corresponds r_i $ for some node $\mathcal{T}_i$}
            \State \textbf{return} \Call{Return}{$\mathcal{T}_i,\pmb{s}$}
        \ElsIf{$n \corresponds d_i$ for some leaf node $L_i$}
            \State \textbf{return} \Call{Decision}{$L_i,\pmb{s}$}
        \Else
            \If{$p \corresponds x_k \in \mathcal{X} ~\forall p \in \mathcal{P}$}
                \State \textbf{return} \Call{ExternalState}{$X^{(t)},\pmb{s}$}
            \Else
                \State \textbf{return} \Call{InternalState}{$X^{(t)},\pmb{s},\pmb{v}$}\label{algl:function_end}
            \EndIf
        \EndIf
    \EndFunction

    \Function{SetInputState}{$n,\pmb{v},\mathcal{P}$}\funclabel{algf:set_input_state} \label{algl:set_input_state}
        \State Let $\pmb{s}$ be an assignment of variables in $\mathcal{X}$
        \For{$p\in\mathcal{P}$}
            \If{$p \corresponds x_k \in \mathcal{X}$}
                \State Set the value of $x_k \in \pmb{s}$ to the value of $p \in \pmb{v}$ 
            \EndIf
        \EndFor
        \State \textbf{return} $\pmb{s}$ \label{algl:set_input_state_end}
    \EndFunction   
    \end{algorithmic}
\end{algorithm}

\begin{algorithm}
    \caption{Defining $\mathcal{F}$ for execution nodes $E_i$}\label{alg:execution}
    \begin{algorithmic}[1]
        \Function{Execution}{$\mathcal{T}_i$}\funclabel{algf:execution} \label{algl:execution}
            \If{$\mathcal{T}_i$ is the root of the tree}
                \State \textbf{return} $true$
            \ElsIf{$\mathcal{T}_i$ is the left-most child of $Pa(\mathcal{T}_i)$}
                \State $\mathcal{T}_p \gets Pa(\mathcal{T}_i)$
                \State \textbf{return} $E_p$
            \Else
                \State Let $\mathcal{T}_j$ be the sibling of $\mathcal{T}_i$ immediately to its left
                \If{$Pa(\mathcal{T}_i)$ is a sequence node}
                    \State \textbf{return} $true$ if $r_j=\Success$, else $false$
                \ElsIf{$Pa(\mathcal{T}_i)$ is a fallback node}
                    \State \textbf{return} $true$ if $r_j=\Failure$, else $false$ \label{algl:execution_end}
                \EndIf
            \EndIf
        \EndFunction
    \end{algorithmic}
\end{algorithm}

\begin{algorithm}
    \caption{Defining $\mathcal{F}$ for return status nodes $r_i$}\label{alg:return}
    \begin{algorithmic}[1]
        \Function{Return}{$\mathcal{T}_i,\pmb{s}$}\funclabel{algf:return} \label{algl:return}
            \If{$E_i$ is $false$}
                \State \textbf{return} $\Invalid$
            \EndIf
            \If{$\mathcal{T}_i$ is an action node}
                \State \textbf{return} $r_i(d_i,\pmb{s})$
            \ElsIf{$\mathcal{T}_i$ is a condition node}
                \State \textbf{return} $r_i(\pmb{s})$
            \ElsIf{$\mathcal{T}_i$ is a sequence node}
                \For{$\mathcal{T}_c \in Ch(\mathcal{T}_i)$}
                    \If{$r_c \in \{\Running,\Failure,\Invalid\}$}
                        \State \textbf{return} $r_c$
                    \EndIf
                    \State \textbf{return} $\Success$
                \EndFor
            \ElsIf{$\mathcal{T}_i$ is a fallback node}
                \For{$\mathcal{T}_c \in Ch(\mathcal{T}_i)$}
                    \If{$r_c \in \{\Running,\Success,\Invalid\}$}
                        \State \textbf{return} $r_c$
                    \EndIf
                    \State \textbf{return} $\Failure$
                \EndFor
            \EndIf
        \EndFunction
    \end{algorithmic}
\end{algorithm}

\begin{algorithm}
    \caption{Defining $\mathcal{F}$ for decision nodes $d_i$}\label{alg:decision}
    \begin{algorithmic}[1]
        \Function{Decision}{$L_i,\pmb{s}$}\funclabel{algf:decision} \label{algl:decision}
            \If{$E_i$ is $false$}
                \State \textbf{return} $\varnothing$
            \Else
                \State \textbf{return} $d_i(\pmb{s})$
            \EndIf
        \EndFunction
    \end{algorithmic}
\end{algorithm}

\begin{algorithm}
    \caption{Defining $\mathcal{F}$ for state nodes $X^{(t)}$}\label{alg:state_node}
    \begin{algorithmic}[1]
        \Function{ExternalState}{$X^{(t)},\pmb{s}$}\funclabel{algf:external_state} \label{algl:external_state}
            \State Let $X\in\mathcal{X}, X^{(t)} \corresponds X$
            \State \textbf{return} $\mathcal{F}_{\mathcal{S}X}(\pmb{s})$ \label{algl:external_state_end}
        \EndFunction
    
        \Function{InternalState}{$X^{(t)},\pmb{s},\pmb{v}$}\funclabel{algf:internal_state} \label{algl:internal_state}
            \State $\exists L_i$ s.t. $E_i,d_i$ are parents of $X^{(t)}$ in $\mathcal{G}$
            \If{$E_i$ is $false$}
                \State \textbf{return} $X^{(t-1)}$
            \Else
                \State Let $\mathcal{P} = \{p|p\text{ is a parent of }d_i\text{ in }\mathcal{G}\}$
                \State $\pmb{z} \gets $\Call{SetInputState}{$d_i,\pmb{v},\mathcal{P}$}
                \State $\pmb{z}^* \gets p_i(\pmb{z})$
                \State Let $X\in\mathcal{X}, X^{(t)} \corresponds X$
                \State \textbf{return} value of $X \in \pmb{z}$ \label{algl:internal_state_end}
            \EndIf
        \EndFunction
    \end{algorithmic}
\end{algorithm}

\section{Serial Recall Task}\label{app:serial_recall_task}
\begin{figure}[t]
    \centering
    \includegraphics[width=\linewidth]{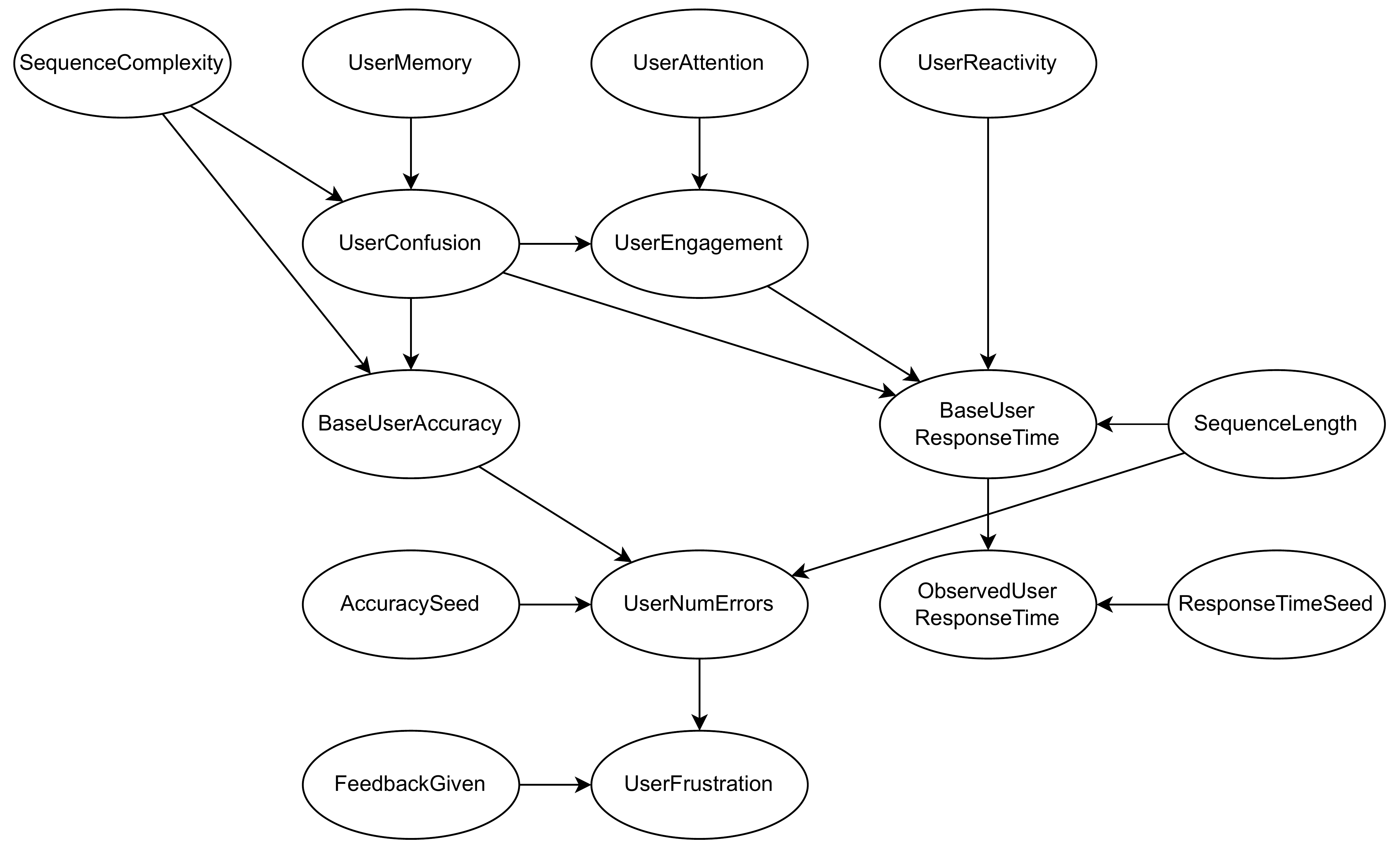}
    \caption{\footnotesize
    The state model $\mathcal{M}_{\mathcal{S}}$ representing the injected domain knowledge in the form of a causal model. Variables listed in Appendix \ref{app:serial_recall_task} but not included in the figure are not causally connected to any other variables (other than through interactions with the BT, as captured by the explanation model).}
    \label{fig:serial_recall_state_model}
\end{figure}

To better contextualise the serial recall problem addressed in Section \ref{sec:evaluation_serial_recall}, we present a diagram of the full behaviour tree, generated by the \textit{py\_trees} library in Figure \ref{fig:serial_recall_bt}. The BT has 33 nodes, 18 of which are leaf nodes. We also provide a depiction of the state model of the environment used in the evaluation in Figure \ref{fig:serial_recall_state_model}. The state model consists of the following variables:

\begin{figure}[t]
    \centering
    \includegraphics[width=\linewidth]{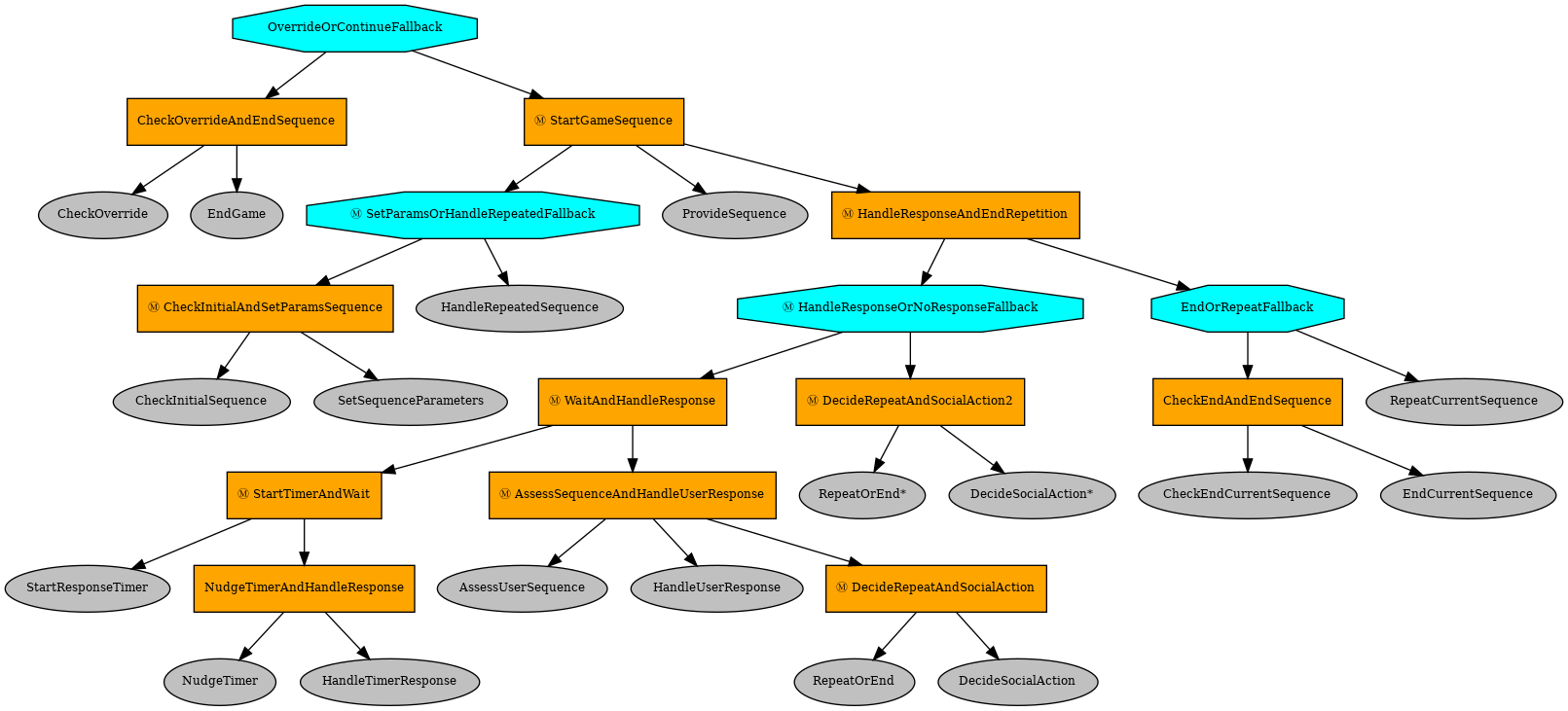}
    \caption{\footnotesize
    The full BT of the serial recall problem presented in Section \ref{sec:evaluation_serial_recall}. The BT has 33 nodes, 18 of which are leaf nodes. Following the visualisation standards of the \textit{py\_trees} library, fallback nodes are represented as light blue octagons, sequence nodes as orange rectangles, and leaf nodes as grey ovals.}
    \label{fig:serial_recall_bt}
\end{figure}

\begin{itemize}
    \item \textit{EndGame} - a boolean flag denoting whether or not the robot should end the entire exercise
    \item \textit{RepeatSequence} - a boolean flag denoting whether or not the robot should repeat the current sequence
    \item \textit{SequenceComplexity} - the number of unique symbols in the sequence. From the set $\{2,3,4\}$
    \item \textit{SequenceLength} - the number of symbols in the sequence. From the set $\{4,5,6,7,8\}$
    \item \textit{NumRepetitions} - the number of times the current sequence has been shown to the user. From the set $\{0,1,2,3\}$
    \item \textit{NumSequences} - the total number of unique sequences shown to the user. From the set $\{0,1,2,3\}$
    \item \textit{SequenceSet} - a boolean flag denoting whether or not the robot has set the current sequence
    \item \textit{ResponseTimerActive} - a boolean flag denoting whether or not the robot is waiting for the user to provide a response
    \item \textit{AttemptedReengageUser} - a boolean flag denoting whether or not the robot has attempted to recapture the attention of the user for the current sequence
    \item \textit{FeedbackGiven} - a boolean flag denoting whether or not the robot has provided feedback (e.g. a hint) to the user for the current repetition of the sequence
    \item \textit{UserResponded} - a boolean flag denoting whether or not the user has provided a response for the current repetition of the sequence
    \item \textit{UserTimeout} - a boolean flag denoting whether or not the user has provided a response within the time limit of 7 seconds 
    \item \textit{CurrentSequence} - a string of \textit{SequenceLength} symbols, composed of \textit{SequenceComplexity} unique symbols. May not be intervened on by the explanation model directly
    \item \textit{UserSequence} - the user's attempt to recall the sequence of symbols. May not be intervened on by the explanation model directly
    \item \textit{AccuracySeed} - a seed used to add noise to the user's accuracy and thus their response. May not be intervened on by the explanation model directly, and is included to satisfy deterministic assumption
    \item \textit{ResponseTimeSeed} - a seed used to add noise to the user's response time. May not be intervened on by the explanation model directly, and is included to satisfy deterministic assumption
    \item \textit{UserMemory} - a variable on the interval $[0,1]$ denoting the ability for the user to recall information
    \item \textit{UserAttention} - a variable on the interval $[0,1]$ denoting the ability for the user to pay attention to events
    \item \textit{UserReactivity} - a variable on the interval $[0,1]$ denoting the reaction speed of the user
    \item \textit{UserConfusion} - a variable on the interval $[0,1]$ denoting how confused a user is in a given situation
    \item \textit{UserEngagement} - a variable on the interval $[0,1]$ denoting how engaged the user is with the task
    \item \textit{UserFrustration} - a variable on the interval $[0,1]$ denoting how frustrated the user is
    \item \textit{BaseUserAccuracy} - a variable on the interval $[0,1]$ denoting the accuracy of the user in recalling the correct sequence
    \item \textit{UserNumErrors} - the number of errors a user makes for the current sequence, on the set $\{0,1,2,3\}$
    \item \textit{BaseUserResponseTime} - the number of seconds, before noise is added, that it takes for a user to respond, on the interval $[0,7]$
    \item \textit{ObservedUserResponseTime} - the number of seconds, after noise is added, that it takes for a user to respond, on the interval $[0,7]$
\end{itemize}

The input, output and action sets (where applicable) are given below for each of the leaf nodes in the BT:

\begin{itemize}
    \item \textit{CheckOverride}
    \begin{itemize}
        \item $\mathcal{X}_i = \{EndGame\}$
    \end{itemize}
    \item \textit{EndGame}
    \begin{itemize}
        \item $\mathcal{A}_i = \{EndGameAction, \varnothing\}$
    \end{itemize}
    \item \textit{CheckInitialSequence}
    \begin{itemize}
        \item $\mathcal{X}_i = \{NumRepetitions\}$
    \end{itemize}
    \item \textit{SetSequenceParameters}
    \begin{itemize}
        \item $\mathcal{X}_i = \{NumSequences, \allowbreak UserMemory, \allowbreak UserAttention, \allowbreak UserReactivity, \allowbreak UserConfusion, \allowbreak UserEngagement, \allowbreak UserNumErrors, \allowbreak UserTimeout, \allowbreak SequenceComplexity, \allowbreak SequenceLength, \allowbreak ObservedUserResponseTime\}$ 
        \item $\mathcal{Y}_i = \{SequenceLength,SequenceComplexity,SequenceSet,NumSequences,CurrentSequence\}$
        \item $\mathcal{A}_i = \{SetSequenceParametersAction(L,C)|L\in\{4,5,6,7,8\}, C\in \{2,3,4\}\}\cup\{\varnothing\}$
    \end{itemize}
    \item \textit{HandleRepeatedSequence}
    \begin{itemize}
        \item $\mathcal{X}_i = \{SequenceSet\}$
        \item $\mathcal{A}_i = \{CrashAction, \varnothing\}$
    \end{itemize}
    \item \textit{ProvideSequence}
    \begin{itemize}
        \item $\mathcal{X}_i = \{SequenceSet, \allowbreak NumRepetitions, \allowbreak SequenceLength, \allowbreak SequenceComplexity, \allowbreak UserMemory, \allowbreak UserAttention, \allowbreak UserReactivity, \allowbreak AccuracySeed, \allowbreak ResponseTimeSeed\}$
        \item $\mathcal{Y}_i = \{NumRepetitions, \allowbreak UserResponded, \allowbreak UserConfusion, \allowbreak UserEngagement, \allowbreak BaseUserAccuracy, \allowbreak UserNumErrors, \allowbreak BaseUserResponseTime, \allowbreak ObservedUserResponseTime, \allowbreak UserSequence, \allowbreak AccuracySeed, \allowbreak ResponseTimeSeed\}$
        \item $\mathcal{A}_i = \{ProvideSequenceAction, \varnothing\}$
    \end{itemize}
    \item \textit{StartResponseTimer}
    \begin{itemize}
        \item $\mathcal{X}_i = \{ResponseTimerActive\}$
        \item $\mathcal{Y}_i = \{ResponseTimerActive,UserResponded,UserTimeout\}$
        \item $\mathcal{A}_i = \{ResetTimerAction, \varnothing\}$
    \end{itemize}
    \item \textit{NudgeTimer}
    \begin{itemize}
        \item $\mathcal{X}_i = \{ObservedUserResponseTime\}$
        \item $\mathcal{Y}_i = \{UserResponded,UserTimeout\}$
        \item $\mathcal{A}_i = \{CheckTimerAction, \varnothing\}$
    \end{itemize}
    \item \textit{HandleTimerResponse}
    \begin{itemize}
        \item $\mathcal{X}_i = \{UserResponded, UserTimeout\}$
        \item $\mathcal{Y}_i = \{ResponseTimerActive\}$
    \end{itemize}
    \item \textit{AssessUserSequence}
    \begin{itemize}
        \item $\mathcal{A}_i = \{AssessSequenceAction, \varnothing\}$
    \end{itemize}
    \item \textit{HandleUserResponse}
    \begin{itemize}
        \item $\mathcal{X}_i = \{UserResponded\}$
    \end{itemize}
    \item \textit{RepeatOrEnd}
    \begin{itemize}
        \item $\mathcal{X}_i = \{NumRepetitions,\allowbreak UserResponded,\allowbreak UserFrustration,\allowbreak  UserEngagement,\allowbreak AttemptedReengageUser,\allowbreak UserNumErrors\}$
        \item $\mathcal{Y}_i = \{RepeatSequence\}$
        \item $\mathcal{A}_i = \{RepeatThisSequenceAction, EndThisSequenceAction, \varnothing\}$
    \end{itemize}
    \item \textit{DecideSocialAction}
    \begin{itemize}
        \item $\mathcal{X}_i = \{UserResponded,\allowbreak RepeatSequence,\allowbreak UserConfusion,\allowbreak UserEngagement,\allowbreak AttemptedReengageUser\}$
        \item $\mathcal{Y}_i = \{AttemptedReengageUser,FeedbackGiven\}$
        \item $\mathcal{A}_i = \{GiveSequenceHintAction,\allowbreak RepeatSequenceSocialAction,\allowbreak EndSequenceSocialAction,\allowbreak RecaptureAttentionAction,\allowbreak \varnothing\}$
    \end{itemize}
    \item \textit{CheckEndCurrentSequence}
    \begin{itemize}
        \item $\mathcal{X}_i = \{RepeatSequence\}$
    \end{itemize}
    \item \textit{EndCurrentSequence}
    \begin{itemize}
        \item $\mathcal{X}_i = \{NumSequences\}$
        \item $\mathcal{Y}_i = \{NumRepetitions,\allowbreak SequenceSet,\allowbreak ResponseTimerActive,\allowbreak AttemptedReengageUser,\allowbreak RepeatSequence,\allowbreak EndGame,\allowbreak FeedbackGiven\}$
        \item $\mathcal{A}_i = \{EndThisSequenceAction, \varnothing\}$
    \end{itemize}
    \item \textit{RepeatCurrentSequence}
    \begin{itemize}
        \item $\mathcal{A}_i = \{RepeatThisSequenceAction, \varnothing\}$
    \end{itemize}
\end{itemize}

\bibliographystyle{plain}
\bibliography{main}

\end{document}